\documentclass{article} %
\usepackage[preprint]{colm2026_conference}

\usepackage{microtype}
\usepackage{hyperref}
\usepackage{url}
\usepackage{booktabs}
\usepackage{graphicx}
\usepackage{amsmath}
\usepackage{amssymb}
\usepackage{subcaption}
\usepackage{placeins}
\usepackage{multirow}
\usepackage[table]{xcolor}
\usepackage{colortbl}

\definecolor{tableblue}{HTML}{49C0EC}
\definecolor{tablegreen}{HTML}{71C7D4}
\definecolor{tablepurple}{HTML}{8C79B4}

\newcommand{\newcell}[1]{%
  \rotatebox{90}{%
    \parbox{2.0cm}{%
      \setlength{\baselineskip}{0.5em}%
      \textbf{\scriptsize{#1}}
    }%
  }%
}

\usepackage{lineno}

\definecolor{darkblue}{rgb}{0, 0, 0.5}
\hypersetup{colorlinks=true, citecolor=darkblue, linkcolor=darkblue, urlcolor=darkblue}

\title{You Only Judge Once: Multi-response Reward Modeling in a Single Forward Pass}

\author{Yinuo Yang, Zixian Ma, Manasi Ganti, Jieyu Zhang \& Ranjay Krishna \\
Paul G. Allen School of Computer Science \& Engineering \\
University of Washington \\
Seattle, WA 98195, USA \\
\texttt{\{yinuoy, zixianma, mganti, jieyuz2, ranjay\}@cs.washington.edu}
}

\newcommand{\imagebench}{MR$^2$Bench-Image}
\newcommand{\videobench}{MR$^2$Bench-Video}

\begin{document}

\ifcolmsubmission
\linenumbers
\fi

\maketitle

\begin{abstract}
    
We present a discriminative multimodal reward model that scores all candidate responses in a single forward pass. Conventional discriminative reward models evaluate each response independently, requiring multiple forward passes, one for each potential response. Our approach concatenates multiple responses with separator tokens and applies cross-entropy over their scalar scores, enabling direct comparative reasoning and efficient $N$-way preference learning. 
The multi-response design also yields up to $N\times$ wall-clock speedup and FLOPs reduction over conventional single-response scoring. To enable $N$-way reward evaluation beyond existing pairwise benchmarks, we construct two new benchmarks: (1) \imagebench{} contains human-annotated rankings over responses from 8 diverse models; (2) \videobench{} is a large-scale video-based reward benchmark derived from 94K crowdsourced pairwise human judgments over video question-answering spanning 19 models, denoised via preference graph ensemble. Both benchmarks provide 4-response evaluation variants sampled from the full rankings.
Built on a 4B vision-language backbone with LoRA fine-tuning and a lightweight MLP value head, our model achieves state-of-the-art results on six multimodal reward benchmarks, including \imagebench{}, \videobench{}, and four other existing benchmarks. Our model outperforms existing larger generative and discriminative reward models. We further demonstrate that our reward model, when used in reinforcement learning with GRPO, produces improved policy models that maintain performance across standard multimodal benchmarks while substantially improving open-ended generation quality, outperforming a single-response discriminative reward model (RM) baseline by a large margin in both training stability and open-ended generation quality.

\end{abstract}

\section{Introduction}
\label{sec:intro}
Reward models are a central component of preference learning for language and vision-language models (VLM). Trained on human preference judgments, they provide scalar signals for response ranking, reranking, test-time selection, and downstream policy optimization in frameworks such as Reinforcement
Learning from Human Feedback (RLHF) with Proximal Policy Optimization (PPO)~\citep{ouyang2022traininglanguagemodelsfollow,stiennon2022learningsummarizehumanfeedback,ziegler2020finetuninglanguagemodelshuman,schulman2017proximalpolicyoptimizationalgorithms}. 
Current multimodal reward models fall into two categories, each with notable limitations. \textit{Generative judges} prompt a large vision-language model to generate a preference verdict via autoregressive decoding~\citep{zheng2023judgingllmasajudgemtbenchchatbot,xiong2025llavacriticlearningevaluatemultimodal}, with variants that produce thinking traces~\citep{zhang2025r1rewardtrainingmultimodalreward} or critiques~\citep{zhang2025mmrlhfstepforwardmultimodal}. 
This reliance on autoregressive text generation incurs significant latency and scales poorly as context length grows.
The canonical implementation of \textit{Discriminative reward models}~\citep{zang2025internlmxcomposer25rewardsimpleeffectivemultimodal,wang2025skyworkvlrewardeffectivereward} avoids text decoding latency by its nature but scores each response in isolation via separate forward passes, preventing the model from directly comparing candidates. 
This design is particularly inefficient for multimodal inputs where image or video context tokens often account for most of the sequence length, as scoring multiple candidates requires repeatedly recomputing the same visual context for each response. 
Therefore, neither paradigm scales gracefully to the $N$-way ranking scenarios that arise naturally in best-of-$N$ sampling and group-based policy optimization~\citep{shao2024deepseekmathpushinglimitsmathematical}.

We propose a simple yet effective alternative: a \textbf{discriminative multimodal reward model that scores all $N$ candidate responses in a single forward pass}. Our approach concatenates the prompt and all candidate responses into one sequence, extracts per-response scalar scores via a lightweight value head, and trains with a cross-entropy loss over the $N$ response scores. Under the causal attention mask, each response attends to all preceding responses, enabling direct comparative reasoning. This design is both more expressive than independent scoring and more efficient than generative decoding and exhaustive pairwise comparison: our model achieves up to $N\times$ wall-clock speedup and FLOPs reduction over the single-response baseline while improving accuracy.

\begin{figure}[t]
\centering
\includegraphics[width=1.0\columnwidth]{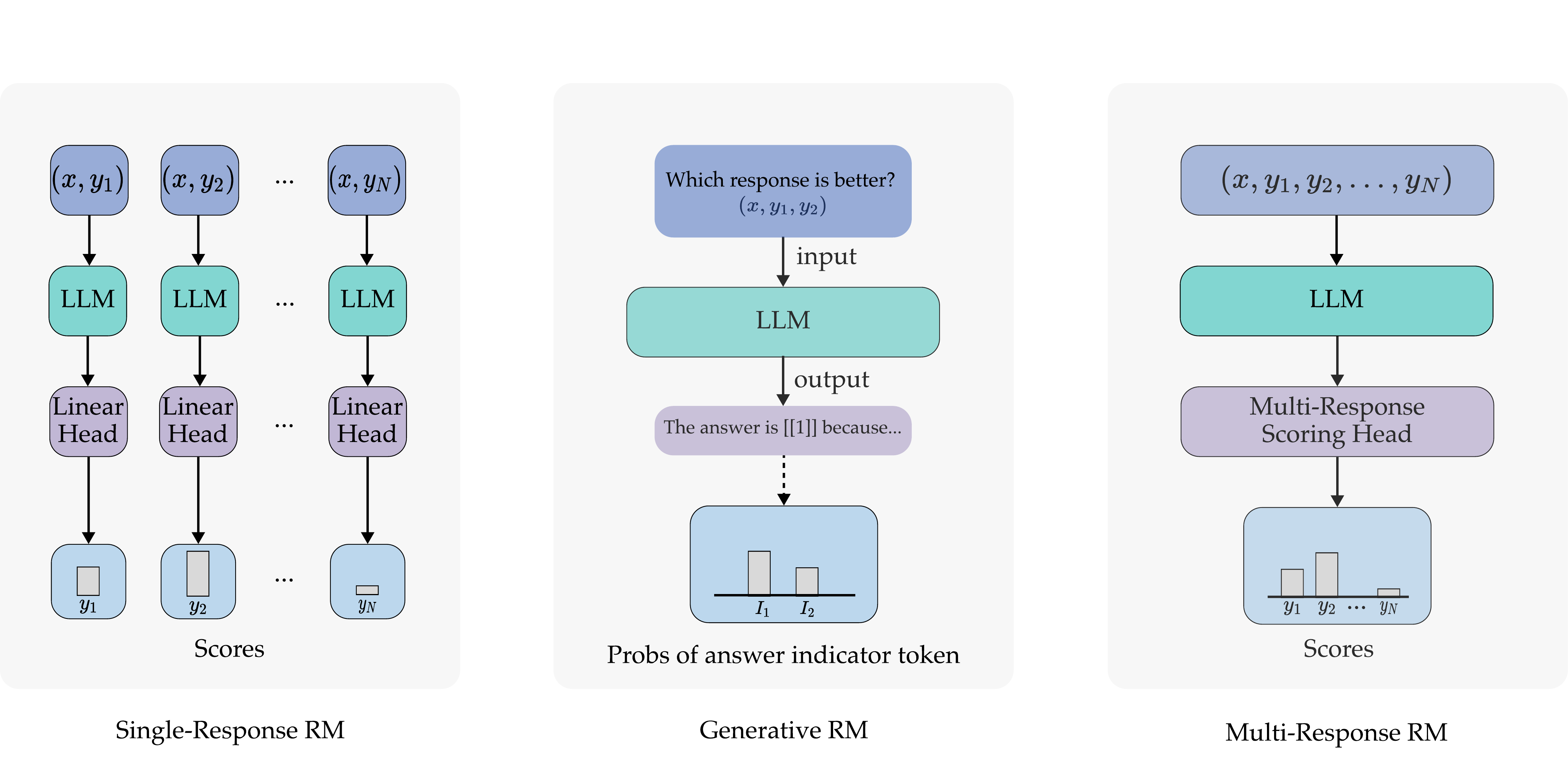}
\caption{\textbf{Comparison of reward model architectures.} \textit{Left:} Single-Response discriminative RM scores each $(x, y_i)$ pair independently via separate forward passes. \textit{Center:} Generative RM prompts a VLM to output a preference distribution $p(I \mid x, y_1, y_2)$ autoregressively. \textit{Right:} Our Multi-Response discriminative RM concatenates all $N$ candidates into a single sequence $(x, y_1, y_2, \ldots, y_N)$ and uses a multi-response scoring head to produce scores for all candidates in one forward pass.}\label{fig:overview}
\vspace{-5mm}
\end{figure}

We also propose the first benchmark for evaluating multimodal reward models on $N$-way comparison for videos. Existing multimodal reward benchmarks~\citep{li2025vlrewardbenchchallengingbenchmarkvisionlanguage,yasunaga2025multimodalrewardbenchholisticevaluation,zhang2025mmrlhfstepforwardmultimodal, zhang2025videorewardbenchcomprehensiveevaluationmultimodal} are limited to pairwise comparisons and offer only limited video coverage. We address this gap with two new \textbf{M}ulti-\textbf{R}esponse \textbf{M}ultimodal \textbf{R}eward Benchmarks: \textbf{\imagebench}, which contains 240 human-annotated rankings over outputs from 8 models across VQA, safety, and visual reasoning, sourced from real user interactions on a VLM playground; and \textbf{\videobench}, which contains 495 video questions with denoised $N$-way rankings over outputs from 19 models, inferred from approximately 94K crowdsourced pairwise judgments.

We build our 4B $N$-way comparison reward model by fine-tuning Molmo2-4B~\citep{clark2026molmo2openweightsdata} with LoRA~\citep{hu2021loralowrankadaptationlarge} on 436K preference samples. Our model achieves state-of-the-art results across all six multimodal reward benchmarks including four image reward benchmarks and two video reward benchmarks, outperforming both larger generative judges and existing discriminative reward models of comparable or greater size.
Additionally, when used as the scoring function in downstream Group Relative Policy Optimization (GRPO)~\citep{shao2024deepseekmathpushinglimitsmathematical}, the model trained with single-response RM is unstable and frequently fails to converge, translating to a substantially weaker downstream improvement. Compared to the single-response RM baseline, our multi-response RM provides a steadily increasing validation reward signal during GRPO training and leads to larger downstream gains.

\section{Related Work}
\label{sec:related_work}

\noindent\textbf{Reward Modeling and Preference Learning.}
Reward models are a core component of preference learning for language models. In the standard RLHF pipeline, a reward model is trained on human preference data, typically with a Bradley--Terry-style~\citep{bradley1952rankanalysis} pairwise objective, and then used to guide downstream policy optimization~\citep{ziegler2020finetuninglanguagemodelshuman,stiennon2022learningsummarizehumanfeedback,ouyang2022traininglanguagemodelsfollow,bai2022traininghelpfulharmlessassistant}. Alternative approaches such as DPO~\citep{rafailov2024directpreferenceoptimizationlanguage} bypass explicit reward modeling. In multimodal settings, early work adapts preference-based alignment to vision-language models, including RLHF-style approaches that train reward models from multimodal human feedback~\citep{sun2023aligninglargemultimodalmodels} and DPO-style approaches that directly optimize VLMs from multimodal preference data or correctional feedback~\citep{yu2024rlhfvtrustworthymllmsbehavior,li2023silkiepreferencedistillationlarge}. More recently, dedicated multimodal reward models have emerged. Discriminative approaches such as IXC-2.5-Reward~\citep{zang2025internlmxcomposer25rewardsimpleeffectivemultimodal} and Skywork-VL-Reward~\citep{wang2025skyworkvlrewardeffectivereward} attach a scalar scoring head to a VLM backbone. Generative approaches such as R1-Reward~\citep{zhang2025r1rewardtrainingmultimodalreward} produce chain-of-thought reasoning before scoring, while MM-RLHF-Reward~\citep{zhang2025mmrlhfstepforwardmultimodal} combines critique generation with scalar scoring. These methods either evaluate each response independently (discriminative) or compare responses pairwise (generative); our method instead processes all $N$ candidates in a single forward pass with cross-entropy, enabling direct comparative reasoning across all candidates simultaneously and more efficient inference. LLM-as-a-judge approaches~\citep{zheng2023judgingllmasajudgemtbenchchatbot} are flexible but computationally expensive at inference time. Our work is complementary, focusing on multi-response scoring efficiency and new $N$-way ranking benchmarks. Concurrent work YOFO~\citep{zhang2026forwardonceefficientcompositional} also pursues single-pass judging, but addresses multi-criterion evaluation for a single input in the recommendation setting, whereas we focus on multi-response ranking for multimodal preference learning.

\noindent\textbf{Reward Benchmarks.}
RewardBench~\citep{lambert2024rewardbenchevaluatingrewardmodels} and RewardBench~2~\citep{malik2025rewardbench2advancingreward} provide standardized evaluation for text-based reward models, with RewardBench~2 introducing more challenging human data and stronger correlation with downstream use. For multimodal reward modeling, benchmarks such as VL-RewardBench~\citep{li2025vlrewardbenchchallengingbenchmarkvisionlanguage}, Multimodal RewardBench~\citep{yasunaga2025multimodalrewardbenchholisticevaluation}, MM-RLHF RewardBench~\citep{zhang2025mmrlhfstepforwardmultimodal}, and VideoRewardBench~\citep{zhang2025videorewardbenchcomprehensiveevaluationmultimodal} substantially broaden evaluation coverage across visual perception, hallucination, reasoning, safety, VQA, and video understanding. However, these multimodal reward benchmarks remain centered on pairwise preference judgments. As a result, they do not directly evaluate a reward model's ability to score multiple candidate responses jointly, which is the relevant setting for best-of-$N$ selection, listwise reranking, and group-based policy optimization. We address this gap with MR$^2$Bench-Image and MR$^2$Bench-Video (Section~\ref{sec:benchmarks}), two multimodal reward benchmarks with explicit $N$-way rankings.

\section{Method}
\label{sec:method}

Conventional discriminative reward models~\citep{ouyang2022traininglanguagemodelsfollow} build on a pretrained language model by appending a linear value head that maps the final hidden state to a scalar reward score $r(x, y)$. Given an input $x$ and a single response $y$, the model processes the concatenation $[x; y]$ in one forward pass to produce the score. To compare $N$ candidate responses, each must be scored in a separate forward pass. Training typically uses the Bradley-Terry (BT) pairwise loss~\citep{bradley1952rankanalysis}:
\begin{equation}
    \mathcal{L}_{\text{BT}} = -\log \sigma\bigl(r(x, y_w) - r(x, y_l)\bigr)
\end{equation}
where $y_w$ and $y_l$ denote the chosen and rejected responses respectively, and $\sigma$ is the sigmoid function.

We present a discriminative multimodal reward model that scores all candidate responses in a single forward pass. Our approach builds on a pretrained vision-language model and introduces three key components: (1) a single-pass multi-response scoring mechanism, (2) a last-token response representation, and (3) a learned value head with cross-entropy training objective. We describe each component below.

\subsection{Single-Pass Multi-Response Scoring}
\label{sec:multi_response}
Our model processes all $N$ candidate responses in a single forward pass. Given a multimodal input $x$ (prompt with optional image or video) and $N$ candidate responses $\{y_1, \ldots, y_N\}$, we concatenate them into one sequence using a special separator token \texttt{<|resp\_sep|>}:
\begin{equation}
    \mathbf{s} = [x; y_1; \texttt{<|resp\_sep|>}; y_2; \texttt{<|resp\_sep|>}; \cdots; y_N]
\end{equation}
The entire sequence is fed through the model once, producing hidden states $\mathbf{H} \in \mathbb{R}^{L \times d}$ over all $L$ tokens. The \texttt{<|resp\_sep|>} token is registered as a special token that always maps to a single unique token ID, providing a reliable anchor for locating response boundaries in the tokenized sequence.

This design offers two advantages. First, efficiency: a single forward pass replaces the $N$ independent passes required by conventional discriminative RMs, yielding up to $N$× computational savings. Second, comparative reasoning: under the causal attention mask, each response attends to all preceding responses and the shared prompt, allowing the model to implicitly contrast candidates rather than scoring them in isolation—a capability absent from independent-scoring approaches.

\subsection{Response Representation}
\label{sec:resp_repr}
For each response $y_i$, the start index $s_i$ is defined as the token immediately after the preceding separator (or the first response token for $y_1$), and the end index $e_i$ is the token immediately before the following separator (or the final token for $y_N$). We extract the hidden state at its last token position $e_i$ to form the response representation:
\begin{equation}
    \mathbf{h}_i = \mathbf{H}_{e_i} \in \mathbb{R}^{d}
\end{equation}
Under the causal attention mask, the last token naturally aggregates information from the entire response, providing a summary representation without requiring additional pooling. We compare this strategy against alternatives (first and last token concatenation, addition, subtraction, and mean pooling) in our ablation study (Table~\ref{tab:ablation}b).

\subsection{Value Head and Training Objective}
\label{sec:value_head}
A two-layer MLP maps each response representation to a scalar reward score:
\begin{equation}
    r_i = \mathbf{w}_2^\top \cdot \sigma(\mathbf{W}_1 \mathbf{h}_i + \mathbf{b}_1) + b_2
\end{equation}
where $\mathbf{W}_1 \in \mathbb{R}^{h \times d}$, $\mathbf{w}_2 \in \mathbb{R}^{h}$, and $\sigma$ is the SiLU activation function, selected from five candidates (ReLU, GeLU, SeLU, Tanh, SiLU) based on our ablation study (Table~\ref{tab:ablation}a). All value head parameters are initialized from $\mathcal{N}(0, 0.01)$ with zero biases.

Given the $N$ scores $\{r_1, \ldots, r_N\}$ and the ground-truth best response index, we minimize a cross-entropy loss:
\begin{equation}
    \mathcal{L} = -\log \frac{\exp(r_{\text{best}})}{\sum_{i=1}^{N} \exp(r_i)}
\end{equation}
When $N{=}2$, this is equivalent to the Bradley-Terry~\citep{bradley1952rankanalysis} pairwise loss, naturally accommodating both pairwise and listwise preference annotations in a unified framework.

\section{Multi-Response Multimodal RewardBench}
\label{sec:benchmarks}

Existing multimodal reward benchmarks (VL-RewardBench~\citep{li2025vlrewardbenchchallengingbenchmarkvisionlanguage}, Multimodal RewardBench~\citep{yasunaga2025multimodalrewardbenchholisticevaluation}, and MM-RLHF RewardBench~\citep{zhang2025mmrlhfstepforwardmultimodal}) are limited to pairwise image comparisons; VideoRewardBench~\citep{zhang2025videorewardbenchcomprehensiveevaluationmultimodal} extends this to video but remains pairwise. None support $N$-way ranking evaluation. We fill this gap by constructing \imagebench{} and \videobench{}, each providing $N$-way human-annotated rankings that enable evaluation of both pairwise and listwise ranking capabilities.

\subsection{\imagebench}
\label{sec:image_benchmark}

We construct \imagebench{} from real user interactions on a VLM playground. Prompts are summarized from user questions and context in dialogues where users consented to data use under the platform's user agreement. We curate 240 prompts paired with uploaded images, spanning three categories: visual question answering (VQA, 80 samples), safety-related queries (80 samples), and visual reasoning (80 samples).

For each prompt-image pair, we generate responses from 8 diverse models: GPT-5, GPT-5~Mini, Claude Sonnet~4.5, Gemini~2.5~Flash, Qwen3-VL-2B, Qwen3-VL-32B, Qwen-7B, and LLaVA-7B~\citep{openai2025gpt5,anthropic2025claudesonnet45,comanici2025gemini,bai2025qwen3vltechnicalreport,qwen,liu2023visualinstructiontuning}. Human annotators rank all eight responses from best to worst, providing a complete ground-truth ordering. From the full 8-response rankings, we construct a 4-response variant by randomly sampling 4 of the 8 responses per sample and preserving their relative ranking order.
\vspace{-3mm}

\subsection{\videobench}
\label{sec:video_benchmark}

We build \videobench{} from human preference annotations over video question-answering responses. We curate 497 questions spanning 489 videos sourced from YouTube Creative Commons and Vimeo, covering diverse video understanding tasks including temporal reasoning, action recognition, and visual detail comprehension.

For each question, pairwise human preference judgments are collected over responses from 19 diverse models spanning proprietary APIs and open-source models of varying scales (full list in Appendix~\ref{app:video_models}), yielding approximately 94K annotations in total (collection details in Appendix~\ref{app:video_collection}).

\noindent\textbf{Preference Graph Denoising.}
\label{sec:pged}
Raw pairwise annotations inevitably contain cyclic inconsistencies due to annotator disagreements. We apply the Preference Graph Ensemble and Denoising (PGED) algorithm~\citep{hu2026acyclicpreferenceevaluationlanguage} to obtain consistent rankings. Per-annotator preference graphs are aggregated into an ensemble graph (57,998 edges), then a greedy cycle removal procedure produces a directed acyclic graph (DAG) with 45,036 edges. Topological sort on the per-question DAG yields consistent rankings, from which we construct a 4-response \videobench{} variant (495 questions after filtering).

\noindent\textbf{Evaluation Metrics.}
\label{sec:bench_metrics}
For both benchmarks, we report \textit{best-of-N accuracy}: whether the model's highest-scored response matches the ground-truth rank-1 response. We report results on the 4-response variants (240 samples for image, 495 samples for video); pairwise accuracy and Kendall's~$\tau$ are reported in Appendix Table~\ref{tab:video_full_metrics}.
\vspace{-3mm}

\section{Experiments}

\label{sec:experiments}

\begin{table*}[t]
\begin{center}
\footnotesize
\setlength{\tabcolsep}{3.5pt}
\resizebox{\textwidth}{!}{%
\begin{tabular}{@{}lccccccccc@{}}
  \toprule
  & & \multicolumn{4}{c}{\textbf{Image}} & \multicolumn{2}{c}{\textbf{Video}} & & \\
  \cmidrule(lr){3-6} \cmidrule(lr){7-8}
  \textbf{Model} & \textbf{Size} & \textbf{VL-RB} & \textbf{MM-RB} & \textbf{MMRLHF} & \textbf{MR$^2$B-I} & \textbf{VRB} & \textbf{MR$^2$B-V} & \textbf{Avg} \\
  \midrule
  \multicolumn{9}{@{}l}{\textit{Proprietary Models$^\dagger$}} \\
  GPT-5~\citep{openai2025gpt5}                & --   & \textbf{75.0} & 64.6 & \textbf{71.8} & \textbf{87.1} & \textbf{68.2} & \textbf{50.1} & \textbf{69.5} \\
  Claude-Sonnet-4.5~\citep{anthropic2025claudesonnet45} & --  & 68.6 & \underline{78.2} & 70.0 & \underline{72.9} & \underline{67.5} & 49.1 & 67.7 \\
  Gemini-2.5-Pro~\citep{comanici2025gemini}    & --   & \underline{70.5} & \textbf{82.4} & \underline{70.6} & 71.2 & 63.2 & \underline{49.7} & \underline{67.9} \\
  \midrule
  \multicolumn{9}{@{}l}{\textit{Open-Source General VLMs$^\dagger$}} \\
  InternVL3-8B~\citep{zhu2025internvl3exploringadvancedtraining}     & 8B   & 56.6 & 66.9 & 69.4 & 55.4 & 57.9 & 40.4 & 57.8 \\
  Qwen2.5-VL-7B~\citep{bai2025qwen25vltechnicalreport}    & 7B   & 66.7 & 62.6 & 77.6 & 52.5 & 55.3 & 44.4 & 59.9 \\
  Qwen3-VL-4B~\citep{bai2025qwen3vltechnicalreport}      & 4B   & 61.4 & 65.9 & \underline{80.0} & 60.8 & \underline{64.9} & \underline{47.9} & 63.5 \\
  Qwen3-VL-8B~\citep{bai2025qwen3vltechnicalreport}      & 8B   & 64.7 & 71.6 & 73.5 & 60.4 & 62.0 & 47.7 & 63.3 \\
  Qwen3-VL-32B~\citep{bai2025qwen3vltechnicalreport}     & 32B  & \underline{67.1} & \textbf{79.0} & 78.8 & 60.8 & \textbf{65.8} & \textbf{49.9} & \textbf{66.9} \\
  Molmo2-4B~\citep{clark2026molmo2openweightsdata}        & 4B   & 59.6 & 61.8 & 73.5 & \underline{61.7} & 58.2 & 43.2 & 59.7 \\
  Molmo2-8B~\citep{clark2026molmo2openweightsdata}        & 8B   & \textbf{68.4} & 66.8 & 68.2 & 60.0 & 57.1 & 42.6 & 60.5 \\
  InternVL3-78B~\citep{zhu2025internvl3exploringadvancedtraining}    & 78B  & 61.9 & \underline{75.7} & \textbf{81.8} & \textbf{65.0} & 58.5 & 47.7 & \underline{65.1} \\
  \midrule
  \multicolumn{9}{@{}l}{\textit{Open-Source Generative Reward Models}} \\
  R1-Reward~\citep{zhang2025r1rewardtrainingmultimodalreward}          & 7B   & \textbf{71.4}$^*$ & \textbf{82.2}$^*$ & \underline{80.6}$^*$ & \textbf{58.8} & \textbf{61.2} & \textbf{44.9} & \textbf{66.5} \\
  MM-RLHF-Reward~\citep{zhang2025mmrlhfstepforwardmultimodal}     & 7B   & \underline{51.0}$^*$ & \underline{67.1}$^*$ & \textbf{85.0}$^*$ & 45.0 & \underline{52.2} & 36.6 & \underline{56.1} \\
  LLaVA-Critic~\citep{xiong2025llavacriticlearningevaluatemultimodal}       & 7B   & 44.0$^*$ & 62.2 & 77.6 & \underline{56.3} & 14.7 & \underline{40.2} & 49.2 \\
  \midrule
  \multicolumn{9}{@{}l}{\textit{Open-Source Discriminative Reward Models}} \\
  Skywork-VL-Reward~\citep{wang2025skyworkvlrewardeffectivereward} & 7B  & 69.0$^*$ & \textbf{74.2} & 72.4 & 52.9 & 62.9 & 46.7 & 63.0 \\
  IXC-2.5-Reward~\citep{zang2025internlmxcomposer25rewardsimpleeffectivemultimodal}   & 7B   & \underline{70.0}$^*$ & 66.6$^*$ & 71.2$^*$ & 55.0 & 57.1 & \underline{48.7} & 61.4 \\
  \textbf{Molmo2-4B Multi-response RM (Ours)}  & \textbf{4B} & \textbf{82.2} & \underline{73.2} & \textbf{92.4} & \textbf{62.5} & \textbf{66.3} & \textbf{50.7} & \textbf{71.2} \\
  \textbf{Qwen3-VL-4B Multi-response RM (Ours)}  & \textbf{4B} & 63.3 & 71.2 & \underline{84.7} & \underline{58.8} & \underline{64.9} & 47.5 & \underline{65.1} \\
  \bottomrule
\end{tabular}%
}
\end{center}
\caption{\textbf{Main results on multimodal reward benchmarks.} Our Molmo2-4B RM (4B) achieves the highest average across all open-source models, outperforming larger generative and discriminative baselines.
\textbf{VL-RB}: VL-RewardBench (macro pairwise acc.);
\textbf{MM-RB}: Multimodal RewardBench (pairwise acc.);
\textbf{MMRLHF}: MM-RLHF RewardBench (pairwise acc.);
\textbf{MR$^2$B-I}: \imagebench{} (best-of-4 acc.);
\textbf{VRB}: VideoRewardBench (macro pairwise acc.);
\textbf{MR$^2$B-V}: \videobench{} (best-of-4 acc.).
$^\dagger$Generative judge. $^*$Paper-reported score.
}\label{tab:main_results}
\end{table*}

\begin{table}[t]
\begin{center}
\small
\setlength{\tabcolsep}{4pt}
\begin{tabular}{@{}llccccccc@{}}
  \toprule
  \textbf{Base Model} & \textbf{Scoring} & \textbf{VL-RB} & \textbf{MM-RB} & \textbf{MMRLHF} & \textbf{MR$^2$B-I} & \textbf{VRB} & \textbf{MR$^2$B-V} & \textbf{Avg} \\
  \midrule
  Molmo2-4B   & Multiple (CE)   & \textbf{62.1} & \textbf{73.8} & \textbf{88.8} & \textbf{52.5} & \textbf{64.3} & \textbf{47.1} & \textbf{64.8} \\
  Molmo2-4B   & Single (BT)     & 57.7 & 61.9 & 64.7 & 41.2 & 60.8 & 37.6 & 54.0 \\
  \midrule
  Qwen3-VL-4B & Multiple (CE)   & 63.6 & 71.2 & 84.7 & \textbf{58.8} & \textbf{64.9} & \textbf{47.5} & \textbf{65.1} \\
  Qwen3-VL-4B & Single (BT)     & \textbf{67.4} & \textbf{73.5} & \textbf{88.8} & 49.6 & 58.1 & 40.6 & 63.0 \\
  \bottomrule
\end{tabular}
\end{center}
\caption{\textbf{Multi-response vs.\ single-response scoring.} Multi-response CE outperforms single-response BT on average, with a large gap on Molmo2-4B (64.8\% vs.\ 54.0\%).}\label{tab:ablation_scoring}
\vspace{-2mm}
\end{table}

\begin{figure}[t]
\centering
\includegraphics[width=0.7\textwidth]{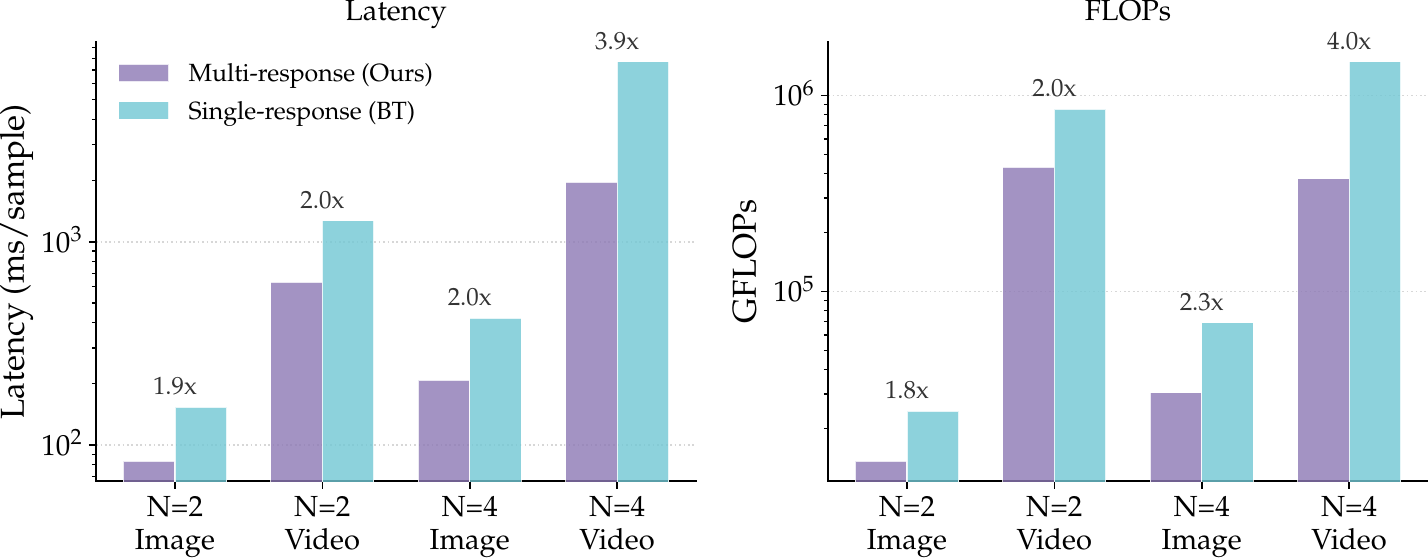}
\caption{\textbf{Inference efficiency of multi-response vs.\ single-response scoring} on Molmo2-4B (single NVIDIA H100 80\,GB GPU). Per-sample latency and FLOPs grouped by $N$ and modality, achieving up to $3.9\times$ latency and $4.0\times$ FLOPs reduction when $N=4$.}\label{fig:latency}
\vspace{2mm}
\centering
\includegraphics[width=0.7\textwidth]{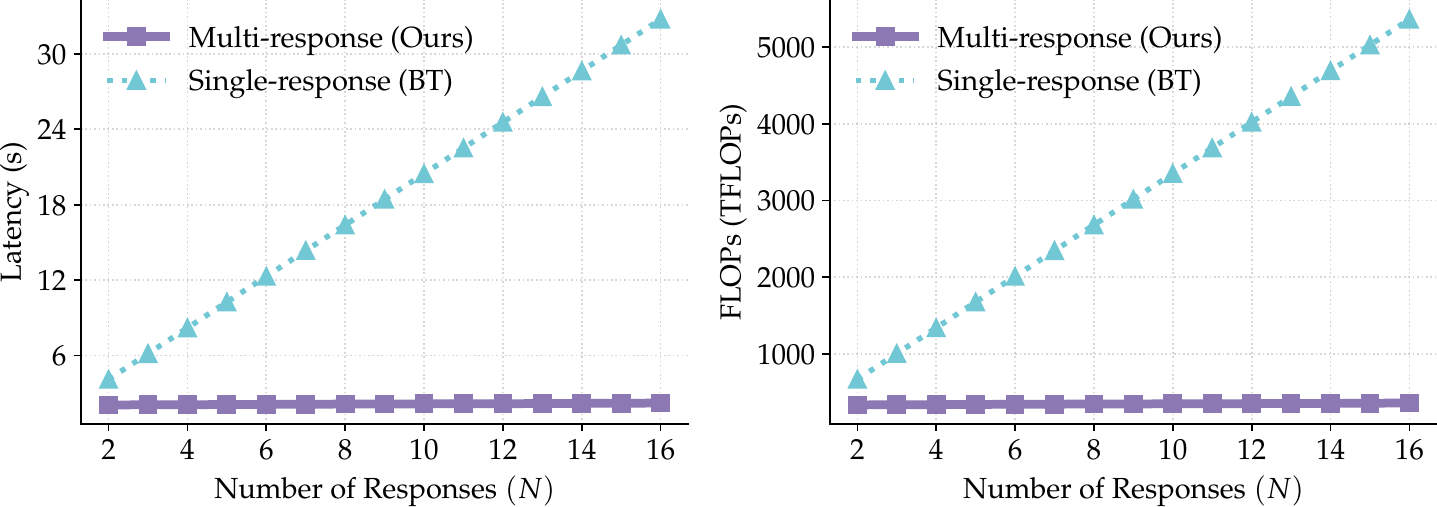}
\caption{\textbf{Efficiency gain scales linearly with $N$.} Plot of latency and FLOPs as $N$ varies. Multi-response cost stays nearly constant while single-response cost grows linearly.}\label{fig:scaling_n}
\vspace{-6mm}
\end{figure}

We evaluate our approach along three axes: (1)~\textit{reward modeling quality}: does our multi-response RM achieve competitive accuracy on multimodal reward benchmarks? (2)~\textit{multi-response vs.\ single-response}: does joint scoring outperform independent scoring in both accuracy and efficiency? (3)~\textit{downstream policy optimization}: can the reward model effectively guide GRPO training? We find that our 4B reward model achieves state-of-the-art results across six multimodal reward benchmarks, that multi-response scoring yields both higher accuracy and up to $N\times$ speedup and FLOPs reduction over single-response scoring, and that GRPO with our multi-response RM substantially improves open-ended generation while preserving standard multi-choice and short answer benchmark performance.

\subsection{Multi-Response Reward Modeling}
\subsubsection{Experimental Setup}
\noindent\textbf{Training Data.}
We curate 436K preference samples from 10 datasets spanning multimodal and text-only sources (Table~\ref{tab:train_datasets_hf_used}; full details in Appendix~\ref{app:training_data}). Notably, 35.1\% of samples contain $N{>}2$ ranked responses, enabling listwise training.

\noindent\textbf{Training Details.}
We build our reward model on top of Molmo2-4B~\citep{clark2026molmo2openweightsdata}, a 4-billion parameter vision-language model with a hidden dimension of $d{=}2560$. The value head uses hidden dimension $h{=}1024$. The vision tower is frozen and the language model is adapted using LoRA~\citep{hu2021loralowrankadaptationlarge} with rank 64, alpha 16, and dropout 0.05. We train for 3 epochs with AdamW (lr = $1 \times 10^{-4}$, no weight decay) and a linear decay schedule without warmup, with effective batch size 64 and maximum sequence length 24{,}576 tokens. During training, we randomly shuffle the order of responses within each sample to prevent the model from developing position bias.

\noindent\textbf{Evaluation Benchmarks.}
We evaluate on four existing multimodal reward benchmarks~\citep{li2025vlrewardbenchchallengingbenchmarkvisionlanguage,yasunaga2025multimodalrewardbenchholisticevaluation,zhang2025mmrlhfstepforwardmultimodal,zhang2025videorewardbenchcomprehensiveevaluationmultimodal} as well as our \imagebench{} and \videobench{}.

\subsubsection{Results}
\noindent\textbf{Benchmark Performance.}
As shown in Table~\ref{tab:main_results}, our Molmo2-4B Multi-response reward model achieves an average of 71.2\% across six benchmarks, outperforming all open-source baselines across generative reward models, discriminative reward models, and general VLMs used as judges. Our Qwen3-VL-4B Multi-response RM achieves 65.1\% average, also competitive with larger baselines, demonstrating that our multi-response approach generalizes across different VLM backbones.

\noindent\textbf{Multi-Response vs.\ Single-Response Scoring.}
\label{sec:multi_vs_single}
We compare multi-response Cross-Entropy (CE) against single-response Bradley-Terry (BT), using the same backbone and training setup on a 73K subset of the full training data. As shown in Table~\ref{tab:ablation_scoring}, on Molmo2-4B, CE achieves substantially higher average accuracy (64.8\% vs.\ 54.0\%). On Qwen3-VL-4B, CE leads on \imagebench{} and \videobench{} while BT is slightly ahead on pairwise benchmarks, resulting in a modest overall gap (65.1\% vs.\ 63.0\%). The gap varies across backbones, suggesting the benefit of cross-response attention interacts with the base model's capabilities.

\noindent\textbf{Inference Speedup.}
Multi-response scoring requires only one forward pass for all $N$ responses, while single-response (BT) requires $N$ passes. As shown in Figure~\ref{fig:latency}, the speedup scales with both $N$ and input length: on $N{=}2$ benchmarks, multi-response scoring achieves ${\sim}1.9\times$ latency and ${\sim}1.8\times$ FLOPs reduction; on $N{=}4$, it reaches up to $3.9\times$ latency and $4.0\times$ FLOPs reduction (video), with image benchmarks at ${\sim}2.0\times$ and ${\sim}2.3\times$ respectively. The speedup approaches $N\times$ when visual tokens dominate the input (as in video), since the shared visual prefix is processed only once; for image benchmarks where response text constitutes a larger fraction of the total sequence, the additional text from concatenating $N$ responses reduces the relative savings. Figure~\ref{fig:scaling_n} confirms this trend: using the source data of \videobench{} (which contains up to 19 model responses per video), we sample 30 videos and vary $N$ from 2 to 16 with our Molmo2-4B CE and BT reward models. Averaged over these samples, multi-response latency stays nearly constant while single-response cost grows linearly. We observe similar efficiency gains with the Qwen3-VL-4B backbone (Appendix~\ref{app:efficiency}).

\subsection{Reinforcement Learning with Multi-response Reward Model}
\label{sec:grpo}
To validate that our multi-response reward model can serve as an effective scoring function for policy optimization, we apply Group Relative Policy Optimization (GRPO)~\citep{shao2024deepseekmathpushinglimitsmathematical} to fine-tune Molmo2-4B using our reward model to score rollout responses.
\subsubsection{Experimental Setup}
We train a GRPO policy model starting from Molmo2-4B on 50K open-ended multimodal prompts, scoring $N{=}4$ rollout responses per prompt with our multi-response RM. The policy uses full fine-tuning (frozen vision tower) for 500 steps with learning rate $1 \times 10^{-5}$ and KL coefficient 0.05. Full training details are provided in Appendix~\ref{app:grpo_setup}.

\subsubsection{Results}
We evaluate across image and video benchmarks, following the Molmo2 evaluation protocol~\citep{clark2026molmo2openweightsdata} (details in Appendix~\ref{app:eval_config}). As shown in Table~\ref{tab:grpo_eval_standard}, GRPO with our multi-response RM preserves performance on all 24 standard multi-choice and short answer multimodal benchmarks. Table~\ref{tab:grpo_eval_openended} shows that it substantially improves open-ended generation: WildVision win rate improves by $+$5.6 (54.6\% $\to$ 60.2\%), LLaVA-Bench by $+$4.6 (92.4 $\to$ 97.0), and MMHal score from 3.98 to 4.25. On video, the policy improves EgoSchema by $+$1.8 and LongVideoBench by $+$1.0 while maintaining other benchmarks.

\begin{figure}[t]
\centering
\includegraphics[width=0.8\textwidth]{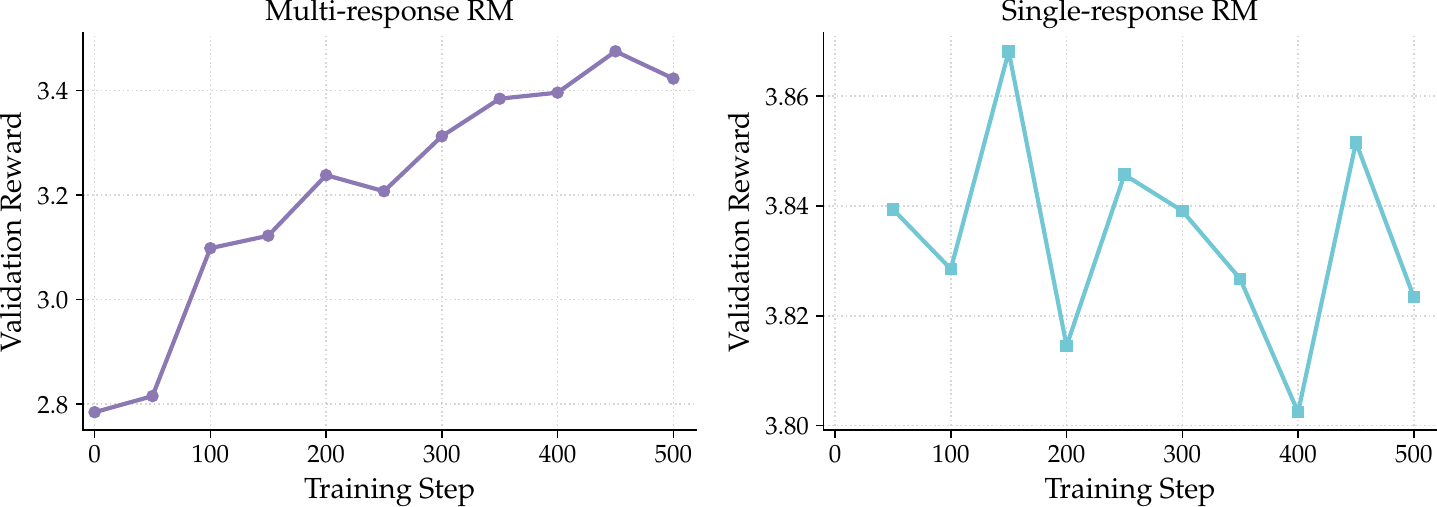}
\caption{\textbf{Validation reward during GRPO training.} The multi-response RM provides a steadily increasing reward signal, while the single-response RM's reward is unstable. The y-axis scales differ because the two reward models produce differently scaled outputs.}\label{fig:grpo_val_reward}
\end{figure}

\noindent\textbf{Multi-response vs.\ single-response RM for GRPO.}
We compare against a single-response BT RM using the same policy setup, reporting the best of several configurations (Appendix~\ref{sec:grpo_single_rm}). As shown in Tables~\ref{tab:grpo_eval_standard} and~\ref{tab:grpo_eval_openended}, the multi-response RM achieves substantially larger open-ended gains (WildVision $+$5.6 vs.\ $+$1.2, LLaVA-W $+$4.6 vs.\ $-$0.8) while better preserving standard benchmarks. We attribute this to the multi-response RM providing a \emph{comparative} reward signal: scoring all $N$ responses jointly directly contrasts candidates rather than assigning independent absolute scores, yielding more informative policy gradients and greater stability. Figure~\ref{fig:grpo_val_reward} confirms this: the multi-response RM's validation reward increases steadily during training, while the single-response RM's remains flat.

\begin{table}[t]
\centering
\small
\setlength{\tabcolsep}{5pt}
\begin{tabular}{@{}lccc@{}}
  \toprule
  \textbf{Model} & \textbf{WildVision} & \textbf{LLaVA-W} & \textbf{MMHal} \\
  \midrule
  Molmo2-4B (base) & 54.6 & 92.4 & 3.98 \\
  \midrule
  + GRPO (Multi-RM) & \textbf{60.2} {\scriptsize\textcolor{tablepurple}{($+$5.6)}} & \textbf{97.0} {\scriptsize\textcolor{tablepurple}{($+$4.6)}} & \textbf{4.25} {\scriptsize\textcolor{tablepurple}{($+$0.27)}} \\
  + GRPO (Single-RM) & 55.8 {\scriptsize\textcolor{tablepurple}{($+$1.2)}} & 91.6 {\scriptsize\textcolor{tablegreen}{($-$0.8)}} & 4.17 {\scriptsize\textcolor{tablepurple}{($+$0.19)}} \\
  \bottomrule
\end{tabular}
\caption{\textbf{GRPO on open-ended generation.} Multi-RM substantially improves three open-ended benchmarks, while Single-RM shows little gains and even hurts LLaVA-Bench.}\label{tab:grpo_eval_openended}
\end{table}

\begin{table*}[!htbp]
\renewcommand{\arraystretch}{0.98}
\centering
\setlength{\tabcolsep}{3pt}
\resizebox{\textwidth}{!}{%
\begin{tabular}{@{}l ccccccccccc cc@{}}
    & \newcell{VQAv2} & \newcell{TextVQA} & \newcell{ChartQA} & \newcell{DocVQA} & \newcell{InfoVQA} & \newcell{AI2D} & \newcell{MMMU} & \newcell{RWQA} & \newcell{MathVista} & \newcell{CountBench} & \newcell{PixMoCount} & \newcell{MuirBench} & \newcell{MMIU} \\
    \midrule
    Molmo2-4B (base)
      & 86.7 & \textbf{84.9} & 86.0 & 92.4 & 78.6
      & 95.6 & \textbf{50.9} & 75.4 & \textbf{56.7} & 94.1
      & 88.1 & \hspace{2pt}\vline\hspace{4pt} 60.5 & 55.5 \\
    \midrule
    Molmo2-4B + GRPO (Multi-RM)
      & \textbf{86.7} & 84.7 & \textbf{86.2} & \textbf{92.5} & \textbf{78.6}
      & \textbf{95.7} & 50.6 & \textbf{75.9} & 56.5 & \textbf{94.3}
      & \textbf{88.3} & \hspace{2pt}\vline\hspace{4pt} \textbf{60.7} & \textbf{55.9} \\
    Molmo2-4B + GRPO (Single-RM)
      & 86.7 & 84.9 & 83.1 & 91.6 & 78.6
      & 95.3 & 50.7 & 75.7 & 56.5 & 94.1
      & 88.3 & \hspace{2pt}\vline\hspace{4pt} 54.2 & 55.5 \\
    \bottomrule
\end{tabular}%
}

{\small \textit{(a) Image standard benchmarks.} Columns: single-image QA $|$ multi-image.}
\vspace{2mm}
\resizebox{\textwidth}{!}{%
\begin{tabular}{@{}l ccccccc ccccc@{}}
    & \newcell{MVBench} & \newcell{TOMATO} & \newcell{MotionB.} & \newcell{TempC.} & \newcell{PercTest} & \newcell{EgoSchema} & \newcell{NextQA} & \newcell{VideoMME} & \newcell{+Sub} & \newcell{LVB+Sub} & \newcell{LVB} \\
    \midrule
    Molmo2-4B (base)
      & 75.1 & 39.9 & 61.8 & 72.8 & \textbf{81.8}
      & 58.6 & 85.6 & \hspace{2pt}\vline\hspace{4pt} 69.1 & 73.7
      & 67.4 & 68.2 \\
    \midrule
    Molmo2-4B + GRPO (Multi-RM)
      & \textbf{75.3} & \textbf{40.5} & 61.4 & \textbf{73.2} & 81.7
      & \textbf{60.4} & \textbf{85.6} & \hspace{2pt}\vline\hspace{4pt} \textbf{69.3} & \textbf{73.7}
      & \textbf{67.5} & \textbf{69.2} \\
    Molmo2-4B + GRPO (Single-RM)
      & 75.2 & 40.0 & \textbf{61.9} & 72.8 & 81.7
      & 58.4 & 85.5 & \hspace{2pt}\vline\hspace{4pt} 69.3 & 73.7
      & 67.3 & 68.2 \\
    \bottomrule
\end{tabular}%
}

\vspace{0.2em}
{\small \textit{(b) Video standard benchmarks.} Columns: short video $|$ long video.}

\caption{\textbf{GRPO on standard benchmarks.} Multi-RM preserves performance across all 24 standard image and video benchmarks, while Single-RM degrades on several.
}\label{tab:grpo_eval_standard}
\vspace{-2mm}
\end{table*}

\subsection{Ablations on Multi-response Reward Modeling}
\label{sec:ablation}

We conduct ablation studies on three design axes using the Molmo2-4B backbone with LoRA-64, lr=$10^{-4}$, 3 epochs, batch size 64, trained on a 73K subset of the full training data, evaluating on all six benchmarks (full results in Appendix Table~\ref{tab:ablation}).

\noindent\textbf{Value head architecture (Table~\ref{tab:ablation}a).}
SiLU achieves the highest average (64.8\%) among five activation functions, outperforming ReLU (64.0\%), GeLU (63.8\%), SeLU (63.2\%), and Tanh (60.5\%). A linear baseline achieves a competitive 64.0\%. 
We adopt SiLU for its balanced performance.
BaseReward~\citep{zhang2025baserewardstrongbaselinemultimodal} arrives at the same finding, reporting that a two-layer MLP with SiLU activation outperforms other reward head designs.

\noindent\textbf{Response representation (Table~\ref{tab:ablation}b).}
Last-token pooling achieves the highest average (64.8\%), followed by mean pooling (64.6\%) and first/last token variants (62.7--63.4\%). This is consistent with the causal attention mechanism, where the last token naturally aggregates information from the entire response.

\noindent\textbf{Loss function (Table~\ref{tab:ablation}c).}
Cross-entropy outperforms Plackett-Luce ranking loss on average (64.8\% vs.\ 63.8\%), suggesting that optimizing for the identity of the best response is more effective than modeling the complete ranking order.

\section{Conclusion}
\label{sec:conclusion}

We introduced a discriminative multimodal reward model that scores all $N$ candidate responses in a single forward pass, achieving up to $N\times$ wall-clock speedup and FLOPs reduction over conventional single-response scoring, and state-of-the-art accuracy across six benchmarks with only 4B parameters. When used as the scoring function for GRPO policy optimization, our multi-response reward model substantially improves open-ended generation quality while preserving standard benchmark performance, and provides a steadily increasing validation reward signal that the single-response baseline lacks. We also constructed \imagebench{} and \videobench{}, two $N$-way ranking benchmarks that fill a gap in multimodal reward evaluation infrastructure. We hope our model and benchmarks facilitate further research on scalable preference evaluation and alignment for multimodal models.

\noindent\textbf{Limitations.}
On \videobench{}, even our best model achieves only 50.7\% best-of-4 accuracy, indicating that video preference evaluation remains challenging. Our experiments evaluate up to $N{=}4$ responses; while the architecture supports arbitrary $N$ (limited only by context length), the scaling behavior at larger $N$ remains unexplored. Additionally, unlike generative judges, our model cannot provide natural language rationales for its preferences, which may limit interpretability in deployment scenarios.

\section*{Acknowledgments}
The project was partially supported by a grant from DSO national laboratories.
The project was also supported by the Qualcomm Innovation Fellowship, OpenAI Superalignment Fellowship, and Apple AI/ML PhD Fellowship.

\section*{Ethics Statement}
\label{sec:ethics}

Our work involves training reward models on human preference data and evaluating them on benchmarks that include safety-related content. The training data includes PKU-SafeRLHF~\citep{ji2025pkusaferlhfmultilevelsafetyalignment}, which contains potentially harmful prompts and responses; we use this data solely to train the reward model to distinguish safe from unsafe responses. \imagebench{} is constructed from user interactions with Molmo-7B~\citep{deitke2024molmopixmoopenweights} on the AI2 Playground; prompts are summarized from user questions and context in dialogues where users consented to data use under the platform's user agreement, and only dialogues retained for at least one month without deletion were used. \imagebench{} includes a safety evaluation category to measure whether reward models can correctly penalize harmful outputs. All human annotations for \videobench{} were collected through a crowdsourcing platform with informed consent, and annotators were compensated at fair market rates. The data was collected as part of the Molmo2 data collection effort~\citep{clark2026molmo2openweightsdata}. The videos used are sourced from YouTube Creative Commons and Vimeo public licenses. We acknowledge that reward models can encode biases present in their training data; users deploying these models for content filtering or policy optimization should validate behavior on their target domains.

\section*{Reproducibility Statement}
\label{sec:reproducibility}

We provide full details to facilitate reproduction of our results. Section~\ref{sec:method} specifies the model architecture, including the value head dimensions ($1024 \times d$), activation function (SiLU), and parameter initialization ($\mathcal{N}(0, 0.01)$). Section~\ref{sec:experiments} details the training configuration: LoRA rank 64, alpha 16, dropout 0.05, learning rate $1 \times 10^{-4}$ with linear decay, 3 epochs, effective batch size 64, and maximum sequence length 24{,}576 tokens. Table~\ref{tab:train_datasets_hf_used} lists all training datasets with their HuggingFace identifiers and exact sample counts. The base models (Molmo2-4B, Qwen3-VL-4B) are publicly available. Appendix~\ref{app:baseline_eval} describes the evaluation protocol for each baseline, and Appendix~\ref{app:eval_config} details the GRPO evaluation configuration. We will release our trained reward model weights and benchmark data (\imagebench{} and \videobench{}) upon publication.

\FloatBarrier
\bibliography{colm2026_conference}

@String(CVPR  = {IEEE Conf. Comput. Vis. Pattern Recog.})

@String(NeurIPS = {Adv. Neural Inform. Process. Syst.})

@String(ICLR  = {Int. Conf. Learn. Represent.})

@String(CVPR  = {CVPR})

@String(NeurIPS = {NeurIPS})

@String(ICLR  = {ICLR})

@misc{zhang2025mmrlhfstepforwardmultimodal,
      title={MM-RLHF: The Next Step Forward in Multimodal LLM Alignment}, 
      author={Yi-Fan Zhang and Tao Yu and Haochen Tian and Chaoyou Fu and Peiyan Li and Jianshu Zeng and Wulin Xie and Yang Shi and Huanyu Zhang and Junkang Wu and Xue Wang and Yibo Hu and Bin Wen and Fan Yang and Zhang Zhang and Tingting Gao and Di Zhang and Liang Wang and Rong Jin and Tieniu Tan},
      year={2025},
      eprint={2502.10391},
      archivePrefix={arXiv},
      primaryClass={cs.CL},
      url={https://arxiv.org/abs/2502.10391}, 
}

@misc{xiong2025llavacriticlearningevaluatemultimodal,
      title={LLaVA-Critic: Learning to Evaluate Multimodal Models}, 
      author={Tianyi Xiong and Xiyao Wang and Dong Guo and Qinghao Ye and Haoqi Fan and Quanquan Gu and Heng Huang and Chunyuan Li},
      year={2025},
      eprint={2410.02712},
      archivePrefix={arXiv},
      primaryClass={cs.CV},
      url={https://arxiv.org/abs/2410.02712}, 
}

@misc{yu2025rlaifvopensourceaifeedback,
      title={RLAIF-V: Open-Source AI Feedback Leads to Super GPT-4V Trustworthiness}, 
      author={Tianyu Yu and Haoye Zhang and Qiming Li and Qixin Xu and Yuan Yao and Da Chen and Xiaoman Lu and Ganqu Cui and Yunkai Dang and Taiwen He and Xiaocheng Feng and Jun Song and Bo Zheng and Zhiyuan Liu and Tat-Seng Chua and Maosong Sun},
      year={2025},
      eprint={2405.17220},
      archivePrefix={arXiv},
      primaryClass={cs.CL},
      url={https://arxiv.org/abs/2405.17220}, 
}

@misc{li2024vlfeedbacklargescaleaifeedback,
      title={VLFeedback: A Large-Scale AI Feedback Dataset for Large Vision-Language Models Alignment}, 
      author={Lei Li and Zhihui Xie and Mukai Li and Shunian Chen and Peiyi Wang and Liang Chen and Yazheng Yang and Benyou Wang and Lingpeng Kong and Qi Liu},
      year={2024},
      eprint={2410.09421},
      archivePrefix={arXiv},
      primaryClass={cs.CV},
      url={https://arxiv.org/abs/2410.09421}, 
}

@misc{zhou2024aligningmodalitiesvisionlarge,
      title={Aligning Modalities in Vision Large Language Models via Preference Fine-tuning}, 
      author={Yiyang Zhou and Chenhang Cui and Rafael Rafailov and Chelsea Finn and Huaxiu Yao},
      year={2024},
      eprint={2402.11411},
      archivePrefix={arXiv},
      primaryClass={cs.LG},
      url={https://arxiv.org/abs/2402.11411}, 
}

@misc{lu2024wildvisionevaluatingvisionlanguagemodels,
      title={WildVision: Evaluating Vision-Language Models in the Wild with Human Preferences}, 
      author={Yujie Lu and Dongfu Jiang and Wenhu Chen and William Yang Wang and Yejin Choi and Bill Yuchen Lin},
      year={2024},
      eprint={2406.11069},
      archivePrefix={arXiv},
      primaryClass={cs.CV},
      url={https://arxiv.org/abs/2406.11069}, 
}

@misc{lambert2025tulu3pushingfrontiers,
      title={Tulu 3: Pushing Frontiers in Open Language Model Post-Training}, 
      author={Nathan Lambert and Jacob Morrison and Valentina Pyatkin and Shengyi Huang and Hamish Ivison and Faeze Brahman and Lester James V. Miranda and Alisa Liu and Nouha Dziri and Shane Lyu and Yuling Gu and Saumya Malik and Victoria Graf and Jena D. Hwang and Jiangjiang Yang and Ronan Le Bras and Oyvind Tafjord and Chris Wilhelm and Luca Soldaini and Noah A. Smith and Yizhong Wang and Pradeep Dasigi and Hannaneh Hajishirzi},
      year={2025},
      eprint={2411.15124},
      archivePrefix={arXiv},
      primaryClass={cs.CL},
      url={https://arxiv.org/abs/2411.15124}, 
}

@misc{liu2024skyworkrewardbagtricksreward,
      title={Skywork-Reward: Bag of Tricks for Reward Modeling in LLMs}, 
      author={Chris Yuhao Liu and Liang Zeng and Jiacai Liu and Rui Yan and Jujie He and Chaojie Wang and Shuicheng Yan and Yang Liu and Yahui Zhou},
      year={2024},
      eprint={2410.18451},
      archivePrefix={arXiv},
      primaryClass={cs.AI},
      url={https://arxiv.org/abs/2410.18451}, 
}

@misc{bai2022traininghelpfulharmlessassistant,
      title={Training a Helpful and Harmless Assistant with Reinforcement Learning from Human Feedback}, 
      author={Yuntao Bai and Andy Jones and Kamal Ndousse and Amanda Askell and Anna Chen and Nova DasSarma and Dawn Drain and Stanislav Fort and Deep Ganguli and Tom Henighan and Nicholas Joseph and Saurav Kadavath and Jackson Kernion and Tom Conerly and Sheer El-Showk and Nelson Elhage and Zac Hatfield-Dodds and Danny Hernandez and Tristan Hume and Scott Johnston and Shauna Kravec and Liane Lovitt and Neel Nanda and Catherine Olsson and Dario Amodei and Tom Brown and Jack Clark and Sam McCandlish and Chris Olah and Ben Mann and Jared Kaplan},
      year={2022},
      eprint={2204.05862},
      archivePrefix={arXiv},
      primaryClass={cs.CL},
      url={https://arxiv.org/abs/2204.05862}, 
}

@misc{starling2023,
    title = {Starling-7B: Improving LLM Helpfulness \& Harmlessness with RLAIF},
    author = {Zhu, Banghua and Frick, Evan and Wu, Tianhao and Zhu, Hanlin and Jiao, Jiantao},
    month = {November},
    year = {2023}
}

@misc{ji2025pkusaferlhfmultilevelsafetyalignment,
      title={PKU-SafeRLHF: Towards Multi-Level Safety Alignment for LLMs with Human Preference}, 
      author={Jiaming Ji and Donghai Hong and Borong Zhang and Boyuan Chen and Juntao Dai and Boren Zheng and Tianyi Qiu and Jiayi Zhou and Kaile Wang and Boxuan Li and Sirui Han and Yike Guo and Yaodong Yang},
      year={2025},
      eprint={2406.15513},
      archivePrefix={arXiv},
      primaryClass={cs.AI},
      url={https://arxiv.org/abs/2406.15513}, 
}

@misc{clark2026molmo2openweightsdata,
      title={Molmo2: Open Weights and Data for Vision-Language Models with Video Understanding and Grounding}, 
      author={Christopher Clark and Jieyu Zhang and Zixian Ma and Jae Sung Park and Mohammadreza Salehi and Rohun Tripathi and Sangho Lee and Zhongzheng Ren and Chris Dongjoo Kim and Yinuo Yang and Vincent Shao and Yue Yang and Weikai Huang and Ziqi Gao and Taira Anderson and Jianrui Zhang and Jitesh Jain and George Stoica and Winson Han and Ali Farhadi and Ranjay Krishna},
      year={2026},
      eprint={2601.10611},
      archivePrefix={arXiv},
      primaryClass={cs.CV},
      url={https://arxiv.org/abs/2601.10611}, 
}

@misc{hu2021loralowrankadaptationlarge,
      title={LoRA: Low-Rank Adaptation of Large Language Models}, 
      author={Edward J. Hu and Yelong Shen and Phillip Wallis and Zeyuan Allen-Zhu and Yuanzhi Li and Shean Wang and Lu Wang and Weizhu Chen},
      year={2021},
      eprint={2106.09685},
      archivePrefix={arXiv},
      primaryClass={cs.CL},
      url={https://arxiv.org/abs/2106.09685}, 
}

@misc{li2025vlrewardbenchchallengingbenchmarkvisionlanguage,
      title={VL-RewardBench: A Challenging Benchmark for Vision-Language Generative Reward Models}, 
      author={Lei Li and Yuancheng Wei and Zhihui Xie and Xuqing Yang and Yifan Song and Peiyi Wang and Chenxin An and Tianyu Liu and Sujian Li and Bill Yuchen Lin and Lingpeng Kong and Qi Liu},
      year={2025},
      eprint={2411.17451},
      archivePrefix={arXiv},
      primaryClass={cs.CV},
      url={https://arxiv.org/abs/2411.17451}, 
}

@misc{yasunaga2025multimodalrewardbenchholisticevaluation,
      title={Multimodal RewardBench: Holistic Evaluation of Reward Models for Vision Language Models}, 
      author={Michihiro Yasunaga and Luke Zettlemoyer and Marjan Ghazvininejad},
      year={2025},
      eprint={2502.14191},
      archivePrefix={arXiv},
      primaryClass={cs.CV},
      url={https://arxiv.org/abs/2502.14191}, 
}

@misc{malik2025rewardbench2advancingreward,
      title={RewardBench 2: Advancing Reward Model Evaluation}, 
      author={Saumya Malik and Valentina Pyatkin and Sander Land and Jacob Morrison and Noah A. Smith and Hannaneh Hajishirzi and Nathan Lambert},
      year={2025},
      eprint={2506.01937},
      archivePrefix={arXiv},
      primaryClass={cs.CL},
      url={https://arxiv.org/abs/2506.01937}, 
}

@article{comanici2025gemini,
  title={Gemini 2.5: Pushing the frontier with advanced reasoning, multimodality, long context, and next generation agentic capabilities},
  author={Comanici, Gheorghe and Bieber, Eric and Schaekermann, Mike and Pasupat, Ice and Sachdeva, Noveen and Dhillon, Inderjit and Blistein, Marcel and Ram, Ori and Zhang, Dan and Rosen, Evan and others},
  journal={arXiv preprint arXiv:2507.06261},
  year={2025}
}

@article{qwen,
  title={Qwen Technical Report},
  author={Jinze Bai and Shuai Bai and Yunfei Chu and Zeyu Cui and Kai Dang and Xiaodong Deng and Yang Fan and Wenbin Ge and Yu Han and Fei Huang and Binyuan Hui and Luo Ji and Mei Li and Junyang Lin and Runji Lin and Dayiheng Liu and Gao Liu and Chengqiang Lu and Keming Lu and Jianxin Ma and Rui Men and Xingzhang Ren and Xuancheng Ren and Chuanqi Tan and Sinan Tan and Jianhong Tu and Peng Wang and Shijie Wang and Wei Wang and Shengguang Wu and Benfeng Xu and Jin Xu and An Yang and Hao Yang and Jian Yang and Shusheng Yang and Yang Yao and Bowen Yu and Hongyi Yuan and Zheng Yuan and Jianwei Zhang and Xingxuan Zhang and Yichang Zhang and Zhenru Zhang and Chang Zhou and Jingren Zhou and Xiaohuan Zhou and Tianhang Zhu},
  journal={arXiv preprint arXiv:2309.16609},
  year={2023}
}

@misc{bai2025qwen25vltechnicalreport,
      title={Qwen2.5-VL Technical Report}, 
      author={Shuai Bai and Keqin Chen and Xuejing Liu and Jialin Wang and Wenbin Ge and Sibo Song and Kai Dang and Peng Wang and Shijie Wang and Jun Tang and Humen Zhong and Yuanzhi Zhu and Mingkun Yang and Zhaohai Li and Jianqiang Wan and Pengfei Wang and Wei Ding and Zheren Fu and Yiheng Xu and Jiabo Ye and Xi Zhang and Tianbao Xie and Zesen Cheng and Hang Zhang and Zhibo Yang and Haiyang Xu and Junyang Lin},
      year={2025},
      eprint={2502.13923},
      archivePrefix={arXiv},
      primaryClass={cs.CV},
      url={https://arxiv.org/abs/2502.13923}, 
}

@misc{wang2025skyworkvlrewardeffectivereward,
      title={Skywork-VL Reward: An Effective Reward Model for Multimodal Understanding and Reasoning}, 
      author={Xiaokun Wang and Peiyu Wang and Jiangbo Pei and Wei Shen and Yi Peng and Yunzhuo Hao and Weijie Qiu and Ai Jian and Tianyidan Xie and Xuchen Song and Yang Liu and Yahui Zhou},
      year={2025},
      eprint={2505.07263},
      archivePrefix={arXiv},
      primaryClass={cs.CV},
      url={https://arxiv.org/abs/2505.07263}, 
}

@misc{zang2025internlmxcomposer25rewardsimpleeffectivemultimodal,
      title={InternLM-XComposer2.5-Reward: A Simple Yet Effective Multi-Modal Reward Model}, 
      author={Yuhang Zang and Xiaoyi Dong and Pan Zhang and Yuhang Cao and Ziyu Liu and Shengyuan Ding and Shenxi Wu and Yubo Ma and Haodong Duan and Wenwei Zhang and Kai Chen and Dahua Lin and Jiaqi Wang},
      year={2025},
      eprint={2501.12368},
      archivePrefix={arXiv},
      primaryClass={cs.CV},
      url={https://arxiv.org/abs/2501.12368}, 
}

@misc{bai2025qwen3vltechnicalreport,
      title={Qwen3-VL Technical Report}, 
      author={Shuai Bai and Yuxuan Cai and Ruizhe Chen and Keqin Chen and Xionghui Chen and Zesen Cheng and Lianghao Deng and Wei Ding and Chang Gao and Chunjiang Ge and Wenbin Ge and Zhifang Guo and Qidong Huang and Jie Huang and Fei Huang and Binyuan Hui and Shutong Jiang and Zhaohai Li and Mingsheng Li and Mei Li and Kaixin Li and Zicheng Lin and Junyang Lin and Xuejing Liu and Jiawei Liu and Chenglong Liu and Yang Liu and Dayiheng Liu and Shixuan Liu and Dunjie Lu and Ruilin Luo and Chenxu Lv and Rui Men and Lingchen Meng and Xuancheng Ren and Xingzhang Ren and Sibo Song and Yuchong Sun and Jun Tang and Jianhong Tu and Jianqiang Wan and Peng Wang and Pengfei Wang and Qiuyue Wang and Yuxuan Wang and Tianbao Xie and Yiheng Xu and Haiyang Xu and Jin Xu and Zhibo Yang and Mingkun Yang and Jianxin Yang and An Yang and Bowen Yu and Fei Zhang and Hang Zhang and Xi Zhang and Bo Zheng and Humen Zhong and Jingren Zhou and Fan Zhou and Jing Zhou and Yuanzhi Zhu and Ke Zhu},
      year={2025},
      eprint={2511.21631},
      archivePrefix={arXiv},
      primaryClass={cs.CV},
      url={https://arxiv.org/abs/2511.21631}, 
}

@misc{bradley1952rankanalysis,
      title={Rank Analysis of Incomplete Block Designs: I. The Method of Paired Comparisons},
      author={Ralph Allan Bradley and Milton E. Terry},
      journal={Biometrika},
      volume={39},
      number={3/4},
      pages={324--345},
      year={1952},
      publisher={Oxford University Press},
}

@article{openai2025gpt5,
      title={{GPT-5} System Card},
      author={OpenAI},
      journal={arXiv preprint arXiv:2601.03267},
      year={2025},
      url={https://arxiv.org/abs/2601.03267},
}

@misc{anthropic2025claudesonnet45,
      title={Claude Sonnet 4.5 System Card},
      author={{Anthropic}},
      year={2025},
      url={https://assets.anthropic.com/m/12f214efcc2f457a/original/Claude-Sonnet-4-5-System-Card.pdf},
}

@misc{zhu2025internvl3exploringadvancedtraining,
      title={InternVL3: Exploring Advanced Training and Test-Time Recipes for Open-Source Multimodal Models}, 
      author={Jinguo Zhu and Weiyun Wang and Zhe Chen and Zhaoyang Liu and Shenglong Ye and Lixin Gu and Hao Tian and Yuchen Duan and Weijie Su and Jie Shao and Zhangwei Gao and Erfei Cui and Xuehui Wang and Yue Cao and Yangzhou Liu and Xingguang Wei and Hongjie Zhang and Haomin Wang and Weiye Xu and Hao Li and Jiahao Wang and Nianchen Deng and Songze Li and Yinan He and Tan Jiang and Jiapeng Luo and Yi Wang and Conghui He and Botian Shi and Xingcheng Zhang and Wenqi Shao and Junjun He and Yingtong Xiong and Wenwen Qu and Peng Sun and Penglong Jiao and Han Lv and Lijun Wu and Kaipeng Zhang and Huipeng Deng and Jiaye Ge and Kai Chen and Limin Wang and Min Dou and Lewei Lu and Xizhou Zhu and Tong Lu and Dahua Lin and Yu Qiao and Jifeng Dai and Wenhai Wang},
      year={2025},
      eprint={2504.10479},
      archivePrefix={arXiv},
      primaryClass={cs.CV},
      url={https://arxiv.org/abs/2504.10479}, 
}

@misc{zhang2025r1rewardtrainingmultimodalreward,
      title={R1-Reward: Training Multimodal Reward Model Through Stable Reinforcement Learning}, 
      author={Yi-Fan Zhang and Xingyu Lu and Xiao Hu and Chaoyou Fu and Bin Wen and Tianke Zhang and Changyi Liu and Kaiyu Jiang and Kaibing Chen and Kaiyu Tang and Haojie Ding and Jiankang Chen and Fan Yang and Zhang Zhang and Tingting Gao and Liang Wang},
      year={2025},
      eprint={2505.02835},
      archivePrefix={arXiv},
      primaryClass={cs.CV},
      url={https://arxiv.org/abs/2505.02835}, 
}

@misc{ouyang2022traininglanguagemodelsfollow,
      title={Training language models to follow instructions with human feedback}, 
      author={Long Ouyang and Jeff Wu and Xu Jiang and Diogo Almeida and Carroll L. Wainwright and Pamela Mishkin and Chong Zhang and Sandhini Agarwal and Katarina Slama and Alex Ray and John Schulman and Jacob Hilton and Fraser Kelton and Luke Miller and Maddie Simens and Amanda Askell and Peter Welinder and Paul Christiano and Jan Leike and Ryan Lowe},
      year={2022},
      eprint={2203.02155},
      archivePrefix={arXiv},
      primaryClass={cs.CL},
      url={https://arxiv.org/abs/2203.02155}, 
}

@misc{stiennon2022learningsummarizehumanfeedback,
      title={Learning to summarize from human feedback}, 
      author={Nisan Stiennon and Long Ouyang and Jeff Wu and Daniel M. Ziegler and Ryan Lowe and Chelsea Voss and Alec Radford and Dario Amodei and Paul Christiano},
      year={2022},
      eprint={2009.01325},
      archivePrefix={arXiv},
      primaryClass={cs.CL},
      url={https://arxiv.org/abs/2009.01325}, 
}

@misc{ziegler2020finetuninglanguagemodelshuman,
      title={Fine-Tuning Language Models from Human Preferences}, 
      author={Daniel M. Ziegler and Nisan Stiennon and Jeffrey Wu and Tom B. Brown and Alec Radford and Dario Amodei and Paul Christiano and Geoffrey Irving},
      year={2020},
      eprint={1909.08593},
      archivePrefix={arXiv},
      primaryClass={cs.CL},
      url={https://arxiv.org/abs/1909.08593}, 
}

@misc{rafailov2024directpreferenceoptimizationlanguage,
      title={Direct Preference Optimization: Your Language Model is Secretly a Reward Model}, 
      author={Rafael Rafailov and Archit Sharma and Eric Mitchell and Stefano Ermon and Christopher D. Manning and Chelsea Finn},
      year={2024},
      eprint={2305.18290},
      archivePrefix={arXiv},
      primaryClass={cs.LG},
      url={https://arxiv.org/abs/2305.18290}, 
}

@misc{lambert2024rewardbenchevaluatingrewardmodels,
      title={RewardBench: Evaluating Reward Models for Language Modeling}, 
      author={Nathan Lambert and Valentina Pyatkin and Jacob Morrison and LJ Miranda and Bill Yuchen Lin and Khyathi Chandu and Nouha Dziri and Sachin Kumar and Tom Zick and Yejin Choi and Noah A. Smith and Hannaneh Hajishirzi},
      year={2024},
      eprint={2403.13787},
      archivePrefix={arXiv},
      primaryClass={cs.LG},
      url={https://arxiv.org/abs/2403.13787}, 
}

@misc{zheng2023judgingllmasajudgemtbenchchatbot,
      title={Judging LLM-as-a-Judge with MT-Bench and Chatbot Arena}, 
      author={Lianmin Zheng and Wei-Lin Chiang and Ying Sheng and Siyuan Zhuang and Zhanghao Wu and Yonghao Zhuang and Zi Lin and Zhuohan Li and Dacheng Li and Eric P. Xing and Hao Zhang and Joseph E. Gonzalez and Ion Stoica},
      year={2023},
      eprint={2306.05685},
      archivePrefix={arXiv},
      primaryClass={cs.CL},
      url={https://arxiv.org/abs/2306.05685}, 
}

@misc{sun2023aligninglargemultimodalmodels,
      title={Aligning Large Multimodal Models with Factually Augmented RLHF}, 
      author={Zhiqing Sun and Sheng Shen and Shengcao Cao and Haotian Liu and Chunyuan Li and Yikang Shen and Chuang Gan and Liang-Yan Gui and Yu-Xiong Wang and Yiming Yang and Kurt Keutzer and Trevor Darrell},
      year={2023},
      eprint={2309.14525},
      archivePrefix={arXiv},
      primaryClass={cs.CV},
      url={https://arxiv.org/abs/2309.14525}, 
}

@misc{li2023silkiepreferencedistillationlarge,
      title={Silkie: Preference Distillation for Large Visual Language Models}, 
      author={Lei Li and Zhihui Xie and Mukai Li and Shunian Chen and Peiyi Wang and Liang Chen and Yazheng Yang and Benyou Wang and Lingpeng Kong},
      year={2023},
      eprint={2312.10665},
      archivePrefix={arXiv},
      primaryClass={cs.CV},
      url={https://arxiv.org/abs/2312.10665}, 
}

@misc{yu2024rlhfvtrustworthymllmsbehavior,
      title={RLHF-V: Towards Trustworthy MLLMs via Behavior Alignment from Fine-grained Correctional Human Feedback}, 
      author={Tianyu Yu and Yuan Yao and Haoye Zhang and Taiwen He and Yifeng Han and Ganqu Cui and Jinyi Hu and Zhiyuan Liu and Hai-Tao Zheng and Maosong Sun and Tat-Seng Chua},
      year={2024},
      eprint={2312.00849},
      archivePrefix={arXiv},
      primaryClass={cs.CL},
      url={https://arxiv.org/abs/2312.00849}, 
}

@misc{shao2024deepseekmathpushinglimitsmathematical,
      title={DeepSeekMath: Pushing the Limits of Mathematical Reasoning in Open Language Models}, 
      author={Zhihong Shao and Peiyi Wang and Qihao Zhu and Runxin Xu and Junxiao Song and Xiao Bi and Haowei Zhang and Mingchuan Zhang and Y. K. Li and Y. Wu and Daya Guo},
      year={2024},
      eprint={2402.03300},
      archivePrefix={arXiv},
      primaryClass={cs.CL},
      url={https://arxiv.org/abs/2402.03300}, 
}

@misc{liu2023visualinstructiontuning,
      title={Visual Instruction Tuning}, 
      author={Haotian Liu and Chunyuan Li and Qingyang Wu and Yong Jae Lee},
      year={2023},
      eprint={2304.08485},
      archivePrefix={arXiv},
      primaryClass={cs.CV},
      url={https://arxiv.org/abs/2304.08485}, 
}

@misc{zhang2025videorewardbenchcomprehensiveevaluationmultimodal,
      title={VideoRewardBench: Comprehensive Evaluation of Multimodal Reward Models for Video Understanding}, 
      author={Zhihong Zhang and Xiaojian Huang and Jin Xu and Zhuodong Luo and Xinzhi Wang and Jiansheng Wei and Xuejin Chen},
      year={2025},
      eprint={2509.00484},
      archivePrefix={arXiv},
      primaryClass={cs.CV},
      url={https://arxiv.org/abs/2509.00484}, 
}

@misc{hu2026acyclicpreferenceevaluationlanguage,
      title={Towards Acyclic Preference Evaluation of Language Models via Multiple Evaluators}, 
      author={Zhengyu Hu and Jieyu Zhang and Zhihan Xiong and Alexander Ratner and Kaize Ding and Ranjay Krishna},
      year={2026},
      eprint={2410.12869},
      archivePrefix={arXiv},
      primaryClass={cs.CL},
      url={https://arxiv.org/abs/2410.12869}, 
}

@inproceedings{chartqa,
    title = "{C}hart{QA}: A Benchmark for Question Answering about Charts with Visual and Logical Reasoning",
    author = "Masry, Ahmed and Long, Do and Tan, Jia Qing and Joty, Shafiq and Hoque, Enamul",
    booktitle = "ACL",
    year = "2022"
}

@inproceedings{mvbench,
  title={Mvbench: A comprehensive multi-modal video understanding benchmark},
  author={Li, Kunchang and Wang, Yali and He, Yinan and Li, Yizhuo and Wang, Yi and Liu, Yi and Wang, Zun and Xu, Jilan and Chen, Guo and Luo, Ping and others},
  booktitle={CVPR},
  year={2024}
}

@inproceedings{tomato,
  title={Tomato: Assessing visual temporal reasoning capabilities in multimodal foundation models},
  author={Shangguan, Ziyao and Li, Chuhan and Ding, Yuxuan and Zheng, Yanan and Zhao, Yilun and Fitzgerald, Tesca and Cohan, Arman},
  booktitle={ICLR},
  year={2025}
}

@inproceedings{egoschema,
  title={EgoSchema: A Diagnostic Benchmark for Very Long-form Video Language Understanding},
  author={Karttikeya Mangalam and Raiymbek Akshulakov and Jitendra Malik},
  booktitle={NeurIPS Track on Datasets and Benchmarks},
  year={2023}
}

@inproceedings{nextqa,
  title={Next-qa: Next phase of question-answering to explaining temporal actions},
  author={Xiao, Junbin and Shang, Xindi and Yao, Angela and Chua, Tat-Seng},
  booktitle={CVPR},
  year={2021}
}

@inproceedings{videomme,
  title={Video-MME: The First-Ever Comprehensive Evaluation Benchmark of Multi-modal LLMs in Video Analysis},
  author={Fu, Chaoyou and Dai, Yuhan and Luo, Yondong and Li, Lei and Ren, Shuhuai and Zhang, Renrui and Wang, Zihan and Zhou, Chenyu and Shen, Yunhang and Zhang, Mengdan and others},
  booktitle={CVPR},
  year={2025}
}

@misc{fang2024mmbenchvideolongformmultishotbenchmark,
      title={MMBench-Video: A Long-Form Multi-Shot Benchmark for Holistic Video Understanding}, 
      author={Xinyu Fang and Kangrui Mao and Haodong Duan and Xiangyu Zhao and Yining Li and Dahua Lin and Kai Chen},
      year={2024},
      eprint={2406.14515},
      archivePrefix={arXiv},
      primaryClass={cs.CV},
      url={https://arxiv.org/abs/2406.14515}, 
}

@misc{huang2025mmoperabenchmarkingopenendedassociation,
      title={MM-OPERA: Benchmarking Open-ended Association Reasoning for Large Vision-Language Models}, 
      author={Zimeng Huang and Jinxin Ke and Xiaoxuan Fan and Yufeng Yang and Yang Liu and Liu Zhonghan and Zedi Wang and Junteng Dai and Haoyi Jiang and Yuyu Zhou and Keze Wang and Ziliang Chen},
      year={2025},
      eprint={2510.26937},
      archivePrefix={arXiv},
      primaryClass={cs.LG},
      url={https://arxiv.org/abs/2510.26937},
}

@misc{zhang2025baserewardstrongbaselinemultimodal,
      title={BaseReward: A Strong Baseline for Multimodal Reward Model}, 
      author={Yi-Fan Zhang and Haihua Yang and Huanyu Zhang and Yang Shi and Zezhou Chen and Haochen Tian and Chaoyou Fu and Haotian Wang and Kai Wu and Bo Cui and Xu Wang and Jianfei Pan and Haotian Wang and Zhang Zhang and Liang Wang},
      year={2025},
      eprint={2509.16127},
      archivePrefix={arXiv},
      primaryClass={cs.CV},
      url={https://arxiv.org/abs/2509.16127},
}

@misc{zhang2025llavavideovideoinstructiontuning,
      title={LLaVA-Video: Video Instruction Tuning With Synthetic Data},
      author={Yuanhan Zhang and Jinming Wu and Wei Li and Bo Li and Zejun Ma and Ziwei Liu and Chunyuan Li},
      year={2025},
      eprint={2410.02713},
      archivePrefix={arXiv},
      primaryClass={cs.CV},
      url={https://arxiv.org/abs/2410.02713},
}

@misc{li2024llavaonevisioneasyvisualtask,
      title={LLaVA-OneVision: Easy Visual Task Transfer}, 
      author={Bo Li and Yuanhan Zhang and Dong Guo and Renrui Zhang and Feng Li and Hao Zhang and Kaichen Zhang and Peiyuan Zhang and Yanwei Li and Ziwei Liu and Chunyuan Li},
      year={2024},
      eprint={2408.03326},
      archivePrefix={arXiv},
      primaryClass={cs.CV},
      url={https://arxiv.org/abs/2408.03326}, 
}

@misc{cho2025perceptionlmopenaccessdatamodels,
      title={PerceptionLM: Open-Access Data and Models for Detailed Visual Understanding},
      author={Jang Hyun Cho and Andrea Madotto and Effrosyni Mavroudi and Triantafyllos Afouras and Tushar Nagarajan and Muhammad Maaz and Yale Song and Tengyu Ma and Shuming Hu and Suyog Jain and Miguel Martin and Huiyu Wang and Hanoona Rasheed and Peize Sun and Po-Yao Huang and Daniel Bolya and Nikhila Ravi and Shashank Jain and Tammy Stark and Shane Moon and Babak Damavandi and Vivian Lee and Andrew Westbury and Salman Khan and Philipp Krähenbühl and Piotr Dollár and Lorenzo Torresani and Kristen Grauman and Christoph Feichtenhofer},
      year={2025},
      eprint={2504.13180},
      archivePrefix={arXiv},
      primaryClass={cs.CV},
      url={https://arxiv.org/abs/2504.13180},
}

@article{yu2025minicpmv45cookingefficient,
      title={MiniCPM-V 4.5: Cooking Efficient MLLMs via Architecture, Data, and Training Recipe},
      author={Tianyu Yu and Zefan Wang and Chongyi Wang and Fuwei Huang and Wenshuo Ma and Zhihui He and Tianchi Cai and Weize Chen and Yuxiang Huang and Yuanqian Zhao and Bokai Xu and Junbo Cui and Yingjing Xu and Liqing Ruan and Luoyuan Zhang and Hanyu Liu and Jingkun Tang and Hongyuan Liu and Qining Guo and Wenhao Hu and Bingxiang He and Jie Zhou and Jie Cai and Ji Qi and Zonghao Guo and Chi Chen and Guoyang Zeng and Yuxuan Li and Ganqu Cui and Ning Ding and Xu Han and Yuan Yao and Zhiyuan Liu and Maosong Sun},
      year={2025},
      journal={arXiv preprint arXiv:2509.18154},
      url={https://arxiv.org/abs/2408.01800},
}

@misc{chen2025eagle25boostinglongcontext,
      title={Eagle 2.5: Boosting Long-Context Post-Training for Frontier Vision-Language Models},
      author={Guo Chen and Zhiqi Li and Shihao Wang and Jindong Jiang and Yicheng Liu and Lidong Lu and De-An Huang and Wonmin Byeon and Matthieu Le and Tuomas Rintamaki and Tyler Poon and Max Ehrlich and Tuomas Rintamaki and Tyler Poon and Tong Lu and Limin Wang and Bryan Catanzaro and Jan Kautz and Andrew Tao and Zhiding Yu and Guilin Liu},
      year={2025},
      eprint={2504.15271},
      archivePrefix={arXiv},
      primaryClass={cs.CV},
      url={https://arxiv.org/abs/2504.15271},
}

@misc{li2025videochatflashhierarchicalcompressionlongcontext,
      title={VideoChat-Flash: Hierarchical Compression for Long-Context Video Modeling},
      author={Xinhao Li and Yi Wang and Jiashuo Yu and Xiangyu Zeng and Yuhan Zhu and Haian Huang and Jianfei Gao and Kunchang Li and Yinan He and Chenting Wang and Yu Qiao and Yali Wang and Limin Wang},
      year={2025},
      eprint={2501.00574},
      archivePrefix={arXiv},
      primaryClass={cs.CV},
      url={https://arxiv.org/abs/2501.00574},
}

@misc{glm2024chatglmfamilylargelanguage,
      title={ChatGLM: A Family of Large Language Models from GLM-130B to GLM-4 All Tools},
      author={Team GLM and : and Aohan Zeng and Bin Xu and Bowen Wang and Chenhui Zhang and Da Yin and Dan Zhang and Diego Rojas and Guanyu Feng and Hanlin Zhao and Hanyu Lai and Hao Yu and Hongning Wang and Jiadai Sun and Jiajie Zhang and Jiale Cheng and Jiayi Gui and Jie Tang and Jing Zhang and Jingyu Sun and Juanzi Li and Lei Zhao and Lindong Wu and Lucen Zhong and Mingdao Liu and Minlie Huang and Peng Zhang and Qinkai Zheng and Rui Lu and Shuaiqi Duan and Shudan Zhang and Shulin Cao and Shuxun Yang and Weng Lam Tam and Wenyi Zhao and Xiao Liu and Xiao Xia and Xiaohan Zhang and Xiaotao Gu and Xin Lv and Xinghan Liu and Xinyi Liu and Xinyue Yang and Xixuan Song and Xunkai Zhang and Yifan An and Yifan Xu and Yilin Niu and Yuantao Yang and Yueyan Li and Yushi Bai and Yuxiao Dong and Zehan Qi and Zhaoyu Wang and Zhen Yang and Zhengxiao Du and Zhenyu Hou and Zihan Wang},
      year={2024},
      eprint={2406.12793},
      archivePrefix={arXiv},
      primaryClass={cs.CL},
      url={https://arxiv.org/abs/2406.12793},
}

@misc{yang2025kwaikeyevl15technical,
      title={Kwai Keye-VL 1.5 Technical Report},
      author={Biao Yang and Bin Wen and Boyang Ding and Changyi Liu and Chenglong Chu and Chengru Song and Chongling Rao and Chuan Yi and Da Li and Dunju Zang and Fan Yang and Guorui Zhou and Guowang Zhang and Han Shen and Hao Peng and Haojie Ding and Hao Wang and Haonan Fan and Hengrui Ju and Jiaming Huang and Jiangxia Cao and Jiankang Chen and Jingyun Hua and Kaibing Chen and Kaiyu Jiang and Kaiyu Tang and Kun Gai and Muhao Wei and Qiang Wang and Ruitao Wang and Sen Na and Shengnan Zhang and Siyang Mao and Sui Huang and Tianke Zhang and Tingting Gao and Wei Chen and Wei Yuan and Xiangyu Wu and Xiao Hu and Xingyu Lu and Yi-Fan Zhang and Yiping Yang and Yulong Chen and Zeyi Lu and Zhenhua Wu and Zhixin Ling and Zhuoran Yang and Ziming Li and Di Xu and Haixuan Gao and Hang Li and Jing Wang and Lejian Ren and Qigen Hu and Qianqian Wang and Shiyao Wang and Xinchen Luo and Yan Li and Yuhang Hu and Zixing Zhang},
      year={2025},
      eprint={2509.01563},
      archivePrefix={arXiv},
      primaryClass={cs.CV},
      url={https://arxiv.org/abs/2509.01563},
}

@misc{deitke2024molmopixmoopenweights,
      title={Molmo and PixMo: Open Weights and Open Data for State-of-the-Art Vision-Language Models}, 
      author={Matt Deitke and Christopher Clark and Sangho Lee and Rohun Tripathi and Yue Yang and Jae Sung Park and Mohammadreza Salehi and Niklas Muennighoff and Kyle Lo and Luca Soldaini and Jiasen Lu and Taira Anderson and Erin Bransom and Kiana Ehsani and Huong Ngo and YenSung Chen and Ajay Patel and Mark Yatskar and Chris Callison-Burch and Andrew Head and Rose Hendrix and Favyen Bastani and Eli VanderBilt and Nathan Lambert and Yvonne Chou and Arnavi Chheda and Jenna Sparks and Sam Skjonsberg and Michael Schmitz and Aaron Sarnat and Byron Bischoff and Pete Walsh and Chris Newell and Piper Wolters and Tanmay Gupta and Kuo-Hao Zeng and Jon Borchardt and Dirk Groeneveld and Crystal Nam and Sophie Lebrecht and Caitlin Wittlif and Carissa Schoenick and Oscar Michel and Ranjay Krishna and Luca Weihs and Noah A. Smith and Hannaneh Hajishirzi and Ross Girshick and Ali Farhadi and Aniruddha Kembhavi},
      year={2024},
      eprint={2409.17146},
      archivePrefix={arXiv},
      primaryClass={cs.CV},
      url={https://arxiv.org/abs/2409.17146}, 
}

@misc{schulman2017proximalpolicyoptimizationalgorithms,
      title={Proximal Policy Optimization Algorithms}, 
      author={John Schulman and Filip Wolski and Prafulla Dhariwal and Alec Radford and Oleg Klimov},
      year={2017},
      eprint={1707.06347},
      archivePrefix={arXiv},
      primaryClass={cs.LG},
      url={https://arxiv.org/abs/1707.06347}, 
}

@misc{zhang2026forwardonceefficientcompositional,
      title={You Only Forward Once: An Efficient Compositional Judging Paradigm}, 
      author={Tianlong Zhang and Hongwei Xue and Shilin Yan and Di Wu and Chen Xu and Guannan Zhang and Yunyun Yang},
      year={2026},
      eprint={2511.16600},
      archivePrefix={arXiv},
      primaryClass={cs.AI},
      url={https://arxiv.org/abs/2511.16600}, 
}
\bibliographystyle{colm2026_conference}

\appendix
\section{Appendix}

\subsection{Ablation Study Results}
\label{app:ablation}
Table~\ref{tab:ablation} reports full ablation results across three design axes: value head architecture, response representation, and loss function. All variants use Molmo2-4B with LoRA rank 64, lr=$10^{-4}$, 3 epochs, batch size 64, trained on a 73K subset of the full training data. The default configuration (MLP with SiLU, last-token pooling, cross-entropy loss) achieves the highest average accuracy (64.8\%) and is used in all main experiments. 

\begin{table*}[h!]
\begin{center}
\small
\setlength{\tabcolsep}{4pt}

\textbf{(a) Value Head Architecture}\\[0.3em]
\begin{tabular}{@{}p{3.2cm}ccccccc@{}}
  \toprule
  \textbf{Value Head} & \textbf{VL-RB} & \textbf{MM-RB} & \textbf{MMRLHF} & \textbf{MR$^2$B-I} & \textbf{VRB} & \textbf{MR$^2$B-V} & \textbf{Avg} \\
  \midrule
  MLP (SiLU)         & \underline{62.1} & 73.8 & 88.8 & \textbf{52.5} & 64.3 & 47.1 & \textbf{64.8} \\
  MLP (SeLU)         & 59.6 & 72.5 & \underline{91.2} & 42.5 & \textbf{66.7} & 46.9 & 63.2 \\
  MLP (ReLU)         & 61.5 & \textbf{74.5} & 88.8 & 47.1 & \underline{65.7} & 46.5 & \underline{64.0} \\
  MLP (GeLU)         & 60.4 & \underline{74.2} & \textbf{91.8} & 43.3 & 65.4 & \underline{47.5} & 63.8 \\
  MLP (Tanh)         & \textbf{69.1} & 70.3 & 73.5 & 47.1 & 54.9 & \textbf{47.9} & 60.5 \\
  Linear             & 62.0 & 73.8 & 88.2 & \underline{51.2} & 63.7 & 45.2 & \underline{64.0} \\
  \bottomrule
\end{tabular}

\vspace{0.3em}

\textbf{(b) Response Representation}\\[0.3em]
\begin{tabular}{@{}p{3.2cm}ccccccc@{}}
  \toprule
  \textbf{Representation} & \textbf{VL-RB} & \textbf{MM-RB} & \textbf{MMRLHF} & \textbf{MR$^2$B-I} & \textbf{VRB} & \textbf{MR$^2$B-V} & \textbf{Avg} \\
  \midrule
  Last token         & \textbf{62.1} & 73.8 & 88.8 & \textbf{52.5} & 64.3 & \textbf{47.1} & \textbf{64.8} \\
  {[}First, last{]}  & 57.7 & 73.3 & \textbf{89.4} & 49.2 & \underline{64.5} & 46.5 & 63.4 \\
  {[}First $+$ last{]}     & \underline{61.7} & \underline{74.1} & 88.2 & \underline{50.0} & 64.4 & 41.2 & 63.3 \\
  {[}First $-$ last{]}     & 61.0 & \textbf{74.3} & 88.8 & 45.8 & 64.2 & 42.0 & 62.7 \\
  Mean pooling       & 60.8 & 71.9 & \textbf{89.4} & 49.6 & \textbf{68.8} & \textbf{47.1} & \underline{64.6} \\
  \bottomrule
\end{tabular}

\vspace{0.3em}

\textbf{(c) Loss Function}\\[0.3em]
\begin{tabular}{@{}p{3.2cm}ccccccc@{}}
  \toprule
  \textbf{Loss} & \textbf{VL-RB} & \textbf{MM-RB} & \textbf{MMRLHF} & \textbf{MR$^2$B-I} & \textbf{VRB} & \textbf{MR$^2$B-V} & \textbf{Avg} \\
  \midrule
  Cross-entropy & \textbf{62.1} & \textbf{73.8} & \textbf{88.8} & \textbf{52.5} & 64.3 & 47.1 & \textbf{64.8} \\
  Plackett-Luce & 55.9 & 72.3 & 88.2 & 51.7 & \textbf{67.5} & \textbf{47.3} & 63.8 \\
  \bottomrule
\end{tabular}

\end{center}
\vspace{-0.5em}
\caption{Ablation studies on three design axes (Section~\ref{sec:ablation}). Default: MLP (SiLU), last-token pooling, cross-entropy loss.}\label{tab:ablation}
\vspace{-1em}
\end{table*}

\subsection{Baseline Evaluation Methodology}
\label{app:baseline_eval}

The baseline reward models in our evaluation employ different scoring mechanisms. We follow each model's official inference protocol and detail them below.

\noindent\textbf{Discriminative reward models (independent scoring).}
Skywork-VL-Reward~\citep{wang2025skyworkvlrewardeffectivereward} and IXC-2.5-Reward~\citep{zang2025internlmxcomposer25rewardsimpleeffectivemultimodal} attach a scalar reward head to a VLM backbone. Each response is scored independently: the prompt and a single response are formatted as a user--assistant conversation, and the reward head extracts a scalar score from the final hidden state. This requires $N$ forward passes per sample.
MM-RLHF-Reward~\citep{zhang2025mmrlhfstepforwardmultimodal} additionally generates a free-form critique before scoring, doubling the per-response cost ($2N$ passes total).

\noindent\textbf{Generative judges (pairwise comparison).}
All generative baselines share a common evaluation protocol: we run all $\binom{N}{2}$ pairwise comparisons and aggregate win counts as pseudo-scores. For our 4-response benchmarks, this requires 6 comparisons per sample. This protocol applies to R1-Reward~\citep{zhang2025r1rewardtrainingmultimodalreward}, LLaVA-Critic~\citep{xiong2025llavacriticlearningevaluatemultimodal}, open-source VLMs (InternVL3~\citep{zhu2025internvl3exploringadvancedtraining}, Qwen2.5-VL~\citep{bai2025qwen25vltechnicalreport}, Qwen3-VL~\citep{bai2025qwen3vltechnicalreport}, Molmo2~\citep{clark2026molmo2openweightsdata}), and proprietary API models (GPT-5~\citep{openai2025gpt5}, Claude Sonnet~4.5~\citep{anthropic2025claudesonnet45}, Gemini~2.5~Pro~\citep{comanici2025gemini}).

The models differ in their generation format:
\begin{itemize}
  \item \textbf{R1-Reward} generates a chain-of-thought analysis in \texttt{<think>} tags followed by a verdict in \texttt{<answer>} tags.
  \item \textbf{LLaVA-Critic} is fine-tuned from LLaVA-OneVision-7B~\citep{li2024llavaonevisioneasyvisualtask} on image-based critique data. For video, we uniformly sample 16 frames as multi-image input. Since it was trained exclusively on image data, its video performance is limited.
  \item \textbf{Open-source VLMs and API models} receive a structured judge prompt and output a \texttt{[[A]]} or \texttt{[[B]]} verdict. Each model uses its native video processing pipeline.
\end{itemize}

An alternative to pairwise aggregation is \emph{direct} best-of-$N$ selection, where the model receives all $N$ responses in a single prompt and directly chooses the best one. Table~\ref{tab:pairwise_vs_direct} compares these two protocols.

\begin{table}[h!]
\begin{center}
\footnotesize
\setlength{\tabcolsep}{4pt}
\begin{tabular}{@{}lccccccc@{}}
  \toprule
  & & \multicolumn{3}{c}{\textbf{MR$^2$B-Image}} & \multicolumn{3}{c}{\textbf{MR$^2$B-Video}} \\
  \cmidrule(lr){3-5} \cmidrule(lr){6-8}
  \textbf{Model} & \textbf{Size} & \textbf{Pairwise} & \textbf{Direct} & \textbf{$\Delta$} & \textbf{Pairwise} & \textbf{Direct} & \textbf{$\Delta$} \\
  \midrule
  \multicolumn{8}{@{}l}{\textit{Proprietary Models}} \\
  GPT-5                  & --   & 87.1 & 87.5 & +0.4 & 50.1 & 50.5 & +0.4 \\
  Claude-Sonnet-4.5      & --   & 72.9 & 72.5 & -0.4 & 49.1 & 49.3 & +0.2 \\
  Gemini-3-Flash         & --   & 72.1 & 70.0 & -2.1 & 52.3 & 52.5 & +0.2 \\
  \midrule
  \multicolumn{8}{@{}l}{\textit{Open-Source General VLMs}} \\
  InternVL3-8B           & 8B   & 55.4 & 52.1 & -3.3 & 40.4 & 40.6 & +0.2 \\
  Qwen2.5-VL-7B          & 7B   & 52.5 & 55.0 & +2.5 & 44.4 & 43.4 & -1.0 \\
  Qwen3-VL-4B            & 4B   & 60.8 & 59.6 & -1.2 & 47.9 & 45.3 & -2.6 \\
  Qwen3-VL-8B            & 8B   & 60.4 & 57.5 & -2.9 & 47.7 & 47.5 & -0.2 \\
  Qwen3-VL-32B           & 32B  & 60.8 & 67.5 & +6.7 & 49.9 & 50.9 & +1.0 \\
  Molmo2-4B              & 4B   & 61.7 & 57.1 & -4.6 & 43.2 & 43.4 & +0.2 \\
  Molmo2-8B              & 8B   & 60.0 & 62.1 & +2.1 & 42.6 & 43.6 & +1.0 \\
  InternVL3-78B          & 78B  & 65.0 & 52.1 & -12.9 & 47.7 & 47.7 & 0.0 \\
  \midrule
  \multicolumn{8}{@{}l}{\textit{Open-Source Generative Reward Models}} \\
  R1-Reward              & 7B   & 58.8 & 40.0 & -18.8 & 44.9 & 36.0 & -8.9 \\
  LLaVA-Critic           & 7B   & 56.3 & 56.7 & +0.4 & 40.2 & 43.2 & +3.0 \\
  \bottomrule
\end{tabular}
\end{center}
\caption{Comparison of pairwise aggregation vs.\ direct best-of-4 selection on \imagebench{} and \videobench{}.
\textbf{Pairwise}: the model evaluates all $\binom{4}{2}=6$ response pairs and selects the response with the highest win count (as used in Table~\ref{tab:main_results}).
\textbf{Direct}: the model receives all 4 responses simultaneously and directly selects the best one.
$\Delta$ = Direct $-$ Pairwise.
}\label{tab:pairwise_vs_direct}
\end{table}

\noindent\textbf{Our model (single-pass multi-response scoring).}
Our model scores all $N$ responses in a single forward pass by concatenating them with separator tokens and extracting per-response scalar scores from the value head. This requires only 1 forward pass per sample regardless of $N$, yielding significant efficiency gains (Section~\ref{sec:method}).

\subsection{Per-Category Benchmark Details}
\label{app:benchmark_details}

Tables~\ref{tab:image_results_detail} and~\ref{tab:video_results_detail} report per-category breakdowns for \imagebench{}, VideoRewardBench, and \videobench{}, complementing the aggregate results in Table~\ref{tab:main_results}.

\begin{table*}[t]
\begin{center}
\footnotesize
\setlength{\tabcolsep}{5pt}
\begin{tabular}{@{}lccccc@{}}
  \toprule
  \textbf{Model} & \textbf{Size} & \textbf{VQA} & \textbf{Reason.} & \textbf{Safety} & \textbf{MR$^2$B-I} \\
  \midrule
  \multicolumn{6}{@{}l}{\textit{Proprietary Models$^\dagger$}} \\
  GPT-5~\citep{openai2025gpt5}                & --   & \textbf{80.0} & \textbf{87.5} & \textbf{93.8} & \textbf{87.1} \\
  Claude-Sonnet-4.5~\citep{anthropic2025claudesonnet45} & --  & \underline{62.5} & 76.2 & \underline{80.0} & \underline{72.9} \\
  Gemini-2.5-Pro~\citep{comanici2025gemini}    & --   & 61.3 & \underline{77.5} & 75.0 & 71.2 \\
  \midrule
  \multicolumn{6}{@{}l}{\textit{Open-Source General VLMs$^\dagger$}} \\
  InternVL3-8B~\citep{zhu2025internvl3exploringadvancedtraining}     & 8B   & 38.8 & 57.5 & 70.0 & 55.4 \\
  Qwen2.5-VL-7B~\citep{bai2025qwen25vltechnicalreport}    & 7B   & 38.8 & 55.0 & 63.8 & 52.5 \\
  Qwen3-VL-4B~\citep{bai2025qwen3vltechnicalreport}      & 4B   & \textbf{48.8} & 60.0 & 73.8 & 60.8 \\
  Qwen3-VL-8B~\citep{bai2025qwen3vltechnicalreport}      & 8B   & 42.5 & \underline{66.2} & 72.5 & 60.4 \\
  Qwen3-VL-32B~\citep{bai2025qwen3vltechnicalreport}     & 32B  & 47.5 & 58.8 & 76.2 & 60.8 \\
  Molmo2-4B~\citep{clark2026molmo2openweightsdata}        & 4B   & \textbf{48.8} & 58.8 & \textbf{77.5} & \underline{61.7} \\
  Molmo2-8B~\citep{clark2026molmo2openweightsdata}        & 8B   & \textbf{48.8} & 57.5 & 73.8 & 60.0 \\
  InternVL3-78B~\citep{zhu2025internvl3exploringadvancedtraining}    & 78B  & \textbf{48.8} & \textbf{68.8} & \textbf{77.5} & \textbf{65.0} \\
  \midrule
  \multicolumn{6}{@{}l}{\textit{Open-Source Generative Reward Models$^\dagger$}} \\
  R1-Reward~\citep{zhang2025r1rewardtrainingmultimodalreward}          & 7B   & \underline{43.8} & \textbf{65.0} & \textbf{67.5} & \textbf{58.8} \\
  MM-RLHF-Reward~\citep{zhang2025mmrlhfstepforwardmultimodal}     & 7B   & 42.5 & 42.5 & 50.0 & 45.0 \\
  LLaVA-Critic~\citep{xiong2025llavacriticlearningevaluatemultimodal}       & 7B   & \textbf{47.5} & \underline{53.8} & \textbf{67.5} & \underline{56.3} \\
  \midrule
  \multicolumn{6}{@{}l}{\textit{Open-Source Discriminative Reward Models}} \\
  Skywork-VL-Reward~\citep{wang2025skyworkvlrewardeffectivereward} & 7B  & 40.0 & 55.0 & 63.8 & 52.9 \\
  IXC-2.5-Reward~\citep{zang2025internlmxcomposer25rewardsimpleeffectivemultimodal}   & 7B   & 42.5 & \textbf{65.0} & 57.5 & 55.0 \\
  \textbf{Molmo2-4B RM (Ours)}  & \textbf{4B} & \textbf{56.2} & 55.0 & \textbf{76.2} & \textbf{62.5} \\
  \textbf{Qwen3-VL-4B RM (Ours)}  & \textbf{4B} & \underline{47.5} & \underline{57.5} & \underline{71.2} & \underline{58.8} \\
  \bottomrule
\end{tabular}
\end{center}
\caption{Per-category results on \textbf{\imagebench} (best-of-4 accuracy, 240 samples: 80 VQA, 80 reasoning, 80 safety, chance = 25\%).
$^\dagger$Generative judge (LLM-as-a-judge).
Best result per category in \textbf{bold}, second best \underline{underlined}.
}\label{tab:image_results_detail}
\end{table*}

\begin{table*}[t]
\begin{center}
\footnotesize
\setlength{\tabcolsep}{3.5pt}
\resizebox{\textwidth}{!}{%
\begin{tabular}{@{}lcccccccc@{}}
  \toprule
  & & \multicolumn{5}{c}{\textbf{VideoRewardBench}} & & \\
  \cmidrule(lr){3-7}
  \textbf{Model} & \textbf{Size} & \textbf{Perc-S} & \textbf{Perc-L} & \textbf{Know.} & \textbf{Reason} & \textbf{Safety} & \textbf{VRB Macro} & \textbf{MR$^2$B-V} \\
  \midrule
  \multicolumn{9}{@{}l}{\textit{Proprietary Models$^\dagger$}} \\
  GPT-5~\citep{openai2025gpt5}                & --   & \underline{57.4} & \underline{68.9} & 67.6 & \underline{62.6} & \textbf{84.6} & \textbf{68.2} & \textbf{50.1} \\
  Claude-Sonnet-4.5~\citep{anthropic2025claudesonnet45} & --  & 55.4 & \textbf{73.9} & \textbf{73.1} & 61.5 & \underline{73.8} & \underline{67.5} & 49.1 \\
  Gemini-2.5-Pro~\citep{comanici2025gemini}    & --   & \textbf{62.7} & 47.7 & \underline{72.7} & \textbf{68.0} & 64.7 & 63.2 & \underline{49.7} \\
  \midrule
  \multicolumn{9}{@{}l}{\textit{Open-Source General VLMs$^\dagger$}} \\
  InternVL3-8B~\citep{zhu2025internvl3exploringadvancedtraining}     & 8B   & 48.1 & 68.2 & 56.3 & \textbf{57.2} & 59.8 & 57.9 & 40.4 \\
  Qwen2.5-VL-7B~\citep{bai2025qwen25vltechnicalreport}    & 7B   & 47.5 & 57.2 & 52.1 & 50.4 & 69.5 & 55.3 & 44.4 \\
  Qwen3-VL-8B~\citep{bai2025qwen3vltechnicalreport}      & 8B   & 51.1 & 67.1 & \underline{58.4} & 54.3 & \textbf{78.9} & \underline{62.0} & \underline{47.7} \\
  Qwen3-VL-32B~\citep{bai2025qwen3vltechnicalreport}     & 32B  & \textbf{60.8} & \textbf{77.7} & \textbf{61.6} & \underline{56.8} & \underline{72.1} & \textbf{65.8} & \textbf{49.9} \\
  Molmo2-4B~\citep{clark2026molmo2openweightsdata}        & 4B   & 47.9 & 65.4 & \underline{58.4} & 55.0 & 64.1 & 58.2 & 43.2 \\
  Molmo2-8B~\citep{clark2026molmo2openweightsdata}        & 8B   & \underline{55.4} & 59.7 & 50.6 & 47.8 & 71.8 & 57.1 & 42.6 \\
  InternVL3-78B~\citep{zhu2025internvl3exploringadvancedtraining}    & 78B  & 47.9 & \underline{69.6} & 57.8 & 51.8 & 65.5 & 58.5 & \underline{47.7} \\
  \midrule
  \multicolumn{9}{@{}l}{\textit{Open-Source Generative Reward Models$^\dagger$}} \\
  R1-Reward~\citep{zhang2025r1rewardtrainingmultimodalreward}          & 7B   & \textbf{52.3} & \textbf{69.6} & \textbf{57.0} & \textbf{55.4} & \textbf{71.5} & \textbf{61.2} & \textbf{44.9} \\
  MM-RLHF-Reward~\citep{zhang2025mmrlhfstepforwardmultimodal}     & 7B   & \underline{37.5} & \underline{61.1} & \underline{45.4} & \underline{52.5} & \underline{64.4} & \underline{52.2} & 36.6 \\
  LLaVA-Critic~\citep{xiong2025llavacriticlearningevaluatemultimodal}       & 7B   & 26.0 & 6.7 & 34.9 & 4.3 & 1.4 & 14.7 & \underline{40.2} \\
  \midrule
  \multicolumn{9}{@{}l}{\textit{Open-Source Discriminative Reward Models}} \\
  Skywork-VL-Reward~\citep{wang2025skyworkvlrewardeffectivereward} & 7B  & 51.1 & \textbf{72.1} & 53.8 & \underline{55.3} & \underline{82.1} & \underline{62.9} & 46.7 \\
  IXC-2.5-Reward~\citep{zang2025internlmxcomposer25rewardsimpleeffectivemultimodal}   & 7B   & 54.0 & \underline{71.0} & 57.1 & 50.7 & 52.7 & 57.1 & \underline{48.7} \\
  \textbf{Molmo2-4B RM (Ours)}  & \textbf{4B} & \underline{56.9} & 67.8 & \textbf{66.4} & \textbf{60.4} & 80.1 & \textbf{66.3} & \textbf{50.7} \\
  \textbf{Qwen3-VL-4B RM (Ours)}  & \textbf{4B} & \textbf{57.9} & 66.8 & \underline{62.9} & 51.4 & \textbf{85.8} & 64.9 & 47.5 \\
  \bottomrule
\end{tabular}%
}
\end{center}
\caption{Results on \textbf{video reward benchmarks}.
\textbf{VideoRewardBench (VRB)}: VideoRewardBench~\citep{zhang2025videorewardbenchcomprehensiveevaluationmultimodal} (pairwise accuracy across five categories, 1,563 pairs, chance = 50\%);
\textbf{MR$^2$B-V}: \videobench{} (best-of-4 accuracy, 495 samples, chance = 25\%).
$^\dagger$Generative judge (LLM-as-a-judge).
Best result per category in \textbf{bold}, second best \underline{underlined}.
}\label{tab:video_results_detail}
\end{table*}

\subsection{GRPO Evaluation Configuration}
\label{app:eval_config}

Table~\ref{tab:eval_config} details the evaluation split, metric, and pipeline used for each benchmark in Section~\ref{sec:grpo}. We follow the Molmo2 technical report~\citep{clark2026molmo2openweightsdata} as closely as possible; deviations are noted below.

\begin{table*}[t]
\begin{center}
\footnotesize
\setlength{\tabcolsep}{3pt}
\resizebox{\textwidth}{!}{%
\begin{tabular}{@{}llllp{5.5cm}@{}}
  \toprule
  \textbf{Category} & \textbf{Benchmark} & \textbf{Split} & \textbf{Metric} & \textbf{Notes} \\
  \midrule
  \multirow{13}{*}{Image Native}
    & VQAv2        & test-standard & VQA score    & Server submission (EvalAI) \\
    & TextVQA      & val        & VQA score       & \\
    & ChartQA      & test       & Relaxed correctness & \\
    & DocVQA       & test       & ANLS            & Server submission (RRC) \\
    & InfoVQA      & test       & ANLS            & Server submission (RRC) \\
    & AI2D         & test       & Accuracy (transparent) & \\
    & MMMU         & val        & Accuracy        & \\
    & RealWorldQA  & test       & Accuracy        & \\
    & MathVista    & testmini   & Accuracy        & \\
    & CountBench   & test       & Per-category avg & \\
    & PixMoCount   & test       & Per-category avg & \\
    & MuirBench    & val        & Accuracy        & \\
    & MMIU         & val        & Accuracy        & \\
  \midrule
  \multirow{3}{*}{Image Open-ended}
    & WildVision   & test       & Win rate (\%)   & GPT-4 judge via lmms-eval \\
    & LLaVA-Bench  & test       & Overall GPT score & GPT-4 judge via lmms-eval \\
    & MMHal        & test       & Avg score (0--6) / Halluc\% & GPT-4 judge via lmms-eval \\
  \midrule
  \multirow{12}{*}{Video Native}
    & MVBench      & test       & Accuracy (EM)   & \\
    & TOMATO       & test       & Accuracy        & \\
    & MotionBench  & val        & Accuracy (EM)   & \\
    & TempCompass  & test       & MCQ accuracy    & MCQ subtask only; caption matching excluded due to scoring bug in upstream code \\
    & PerceptionTest & val      & MC accuracy     & \\
    & EgoSchema    & val (500)  & MC accuracy     & Molmo2 paper reports test/5000 (server submission); server expired, val/500 used \\
    & NextQA       & test       & MC accuracy     & \\
    & VideoMME     & test       & Accuracy        & \\
    & VideoMME+Sub & test       & Accuracy        & \\
    & LongVideoBench+Sub & val  & Accuracy        & \\
    & LVBench      & test       & Accuracy        & \\
    & VideoEvalPro & test       & Accuracy (EM)   & \\
  \midrule
  \multirow{3}{*}{Video Open-ended}
    & MMBench-Video~\citep{fang2024mmbenchvideolongformmultishotbenchmark} & test      & GPT-4 rating (0--3) & GPT-4-turbo judge \\
    & MM-OPERA RIA~\citep{huang2025mmoperabenchmarkingopenendedassociation} & test       & Success rate (\%) & GPT-4 judge \\
    & MM-OPERA ICA~\citep{huang2025mmoperabenchmarkingopenendedassociation} & test       & Success rate (\%) & GPT-4 judge \\
  \bottomrule
\end{tabular}%
}
\end{center}
\caption{Per-benchmark evaluation configuration for GRPO policy evaluation (Section~\ref{sec:grpo}), following the Molmo2 technical report~\citep{clark2026molmo2openweightsdata}.
Deviations: (1)~validation splits used where test sets are unavailable;
(2)~EgoSchema val/500 instead of test/5000;
(3)~video evaluation uses \texttt{decord} (vs.\ \texttt{torchcodec}), \texttt{max\_frames}=376 (vs.\ 384), and 10K subtitle token cap.}\label{tab:eval_config}
\end{table*}

\subsection{Inference Efficiency}
\label{app:efficiency}

\begin{figure}[h]
\centering
\includegraphics[width=\textwidth]{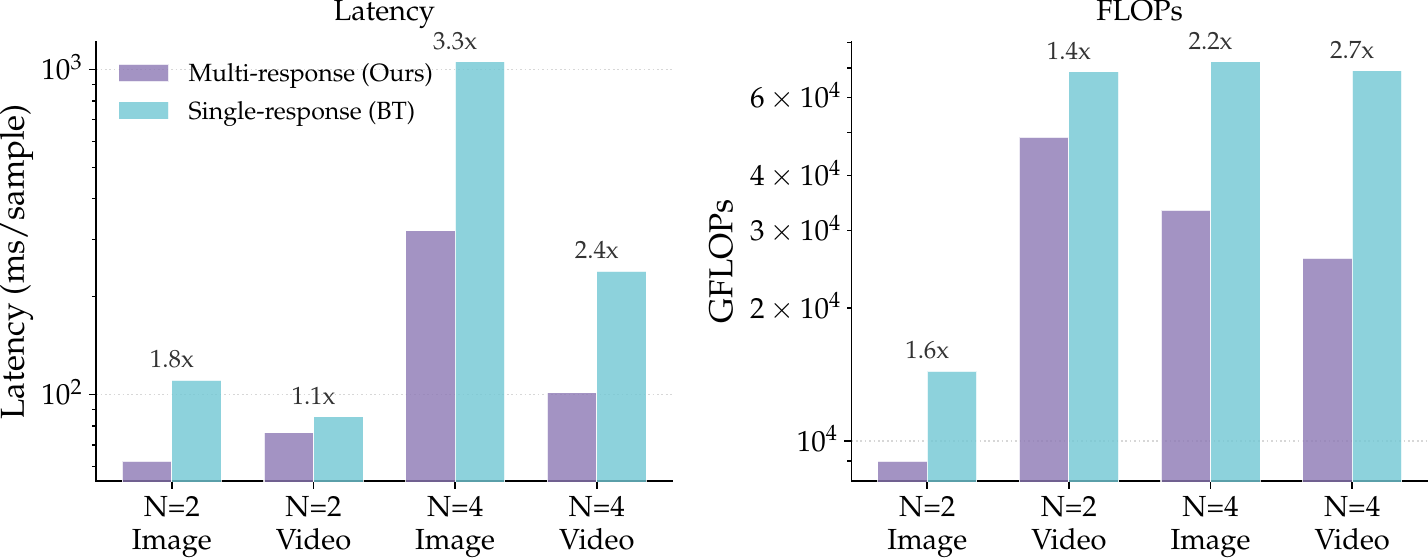}
\caption{Per-sample inference latency (\textbf{left}, ms) and average FLOPs (\textbf{right}) for Qwen3-VL-4B on a single NVIDIA H100 80\,GB GPU. Same grouping as Figure~\ref{fig:latency}.}\label{fig:latency_qwen}
\end{figure}

\noindent\textbf{Multi-response Cross-Entropy (CE) vs.\ single-response Bradley-Terry (BT) on Qwen3-VL.}
Figure~\ref{fig:latency_qwen} shows the latency and FLOPs comparison for Qwen3-VL-4B, complementing the Molmo2-4B results in Figure~\ref{fig:latency}. Notably, Qwen3-VL exhibits higher average inference cost on Image than Video benchmarks, the opposite of Molmo2. This is because Qwen3-VL allocates up to 16{,}384 vision tokens per image (via dynamic resolution) but caps video at 768 tokens per frame, resulting in average input lengths of 4{,}896 tokens for Image vs.\ 2{,}180 for Video across benchmark samples. In contrast, Molmo2 produces shorter sequences for Image (2{,}636 tokens) but much longer ones for Video (12{,}539 tokens). Both the latency and FLOPs panels confirm this pattern, highlighting that inference cost depends not only on modality but also on the model's visual encoding strategy and the input distribution.

\begin{figure}[h]
\centering
\includegraphics[width=\textwidth]{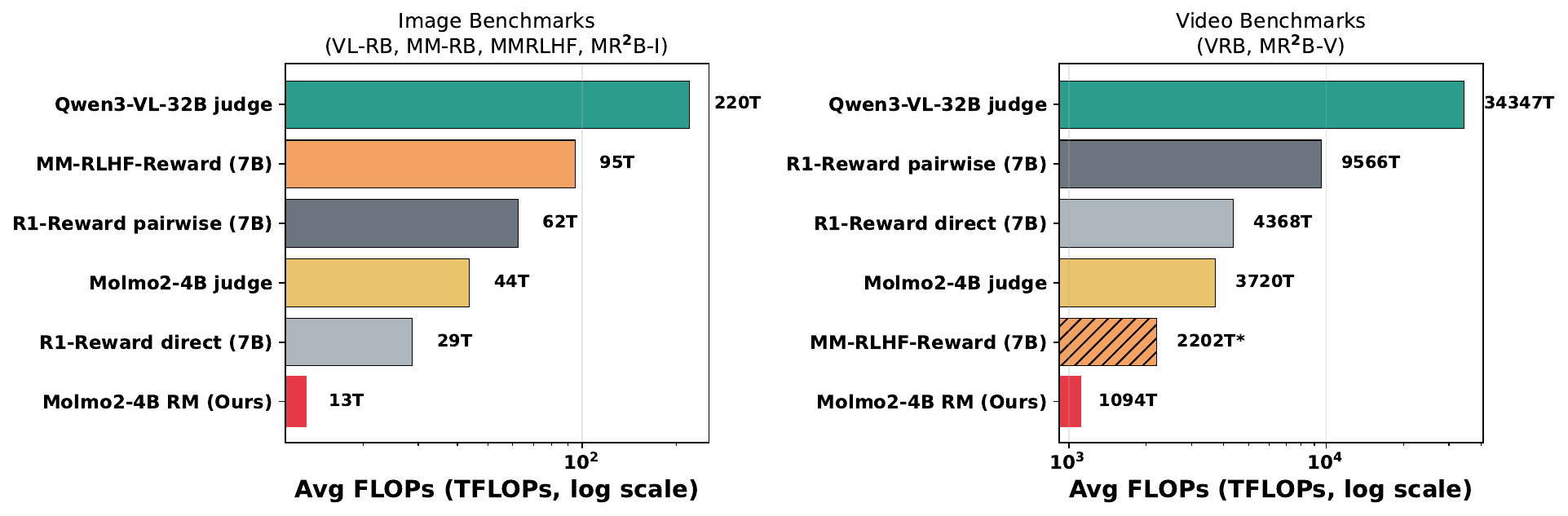}
\caption{Per-sample FLOPs comparison between our Molmo2-4B RM and open-source baselines, averaged over image benchmarks (\textbf{left}: VL-RB, MM-RB, MMRLHF, MR$^2$B-I) and video benchmarks (\textbf{right}: VRB, MR$^2$B-V). MM-RLHF-Reward on video benchmarks ($^*$, hatched) are theoretically estimated due to incomplete automated measurement of LLaVA's internal vision token expansion (see text).}\label{fig:flops_baselines}
\end{figure}

\noindent\textbf{Comparison with baselines.}
\label{app:efficiency_baselines}
Figure~\ref{fig:flops_baselines} compares the per-sample FLOPs of our Molmo2-4B reward model against open-source baselines across image and video benchmarks. FLOPs are measured using PyTorch's \texttt{FlopCounterMode} on a single representative sample per benchmark (FLOPs are deterministic given model architecture and input shape). For each baseline, FLOPs reflect the \emph{total} computation required to rank all $N$ candidate responses in a sample, including all pairwise comparisons or per-response scoring passes.

For MM-RLHF-Reward~\citep{zhang2025mmrlhfstepforwardmultimodal} on video benchmarks (marked with $^*$ and hatching in Figure~\ref{fig:flops_baselines}), \texttt{FlopCounterMode} underestimates FLOPs because LLaVA-OneVision's \texttt{generate()} internally expands a single \texttt{<image>} placeholder into thousands of vision tokens via \texttt{prepare\_inputs\_labels\_for\_multimodal}, and the resulting LLM prefill over these expanded tokens is not fully captured by the flop counter. We therefore estimate video FLOPs theoretically: using the Qwen2-7B architecture (13.1~GFLOPs/token forward), we compute per-response FLOPs as the sum of critique-generation prefill, autoregressive decode, reward-head forward, and vision encoder costs. We calibrate this estimate against the image benchmark, where \texttt{FlopCounterMode} is accurate (vision tokens are few), obtaining a scale factor of $1.37\times$ to account for decode length underestimation. This yields 2{,}937~TFLOPs for \videobench{} (4 responses) and 1{,}468~TFLOPs for VideoRewardBench (2 responses).

Our multi-response scoring requires only a single forward pass regardless of $N$, achieving $2{-}17\times$ lower FLOPs than the most efficient baseline on image benchmarks and remaining competitive on video benchmarks despite using a smaller 4B backbone.

\subsection{Additional Evaluation Metrics}
\label{app:additional_metrics}

Table~\ref{tab:video_full_metrics} reports pairwise accuracy and Kendall's $\tau$ rank correlation alongside best-of-N accuracy for \videobench{}. Our Molmo2-4B RM achieves the highest pairwise accuracy among discriminative reward models, confirming that its ranking quality extends beyond top-1 selection.

\begin{table}[h]
\begin{center}
\footnotesize
\setlength{\tabcolsep}{4pt}
\begin{tabular}{@{}lcccc@{}}
  \toprule
  \textbf{Model} & \textbf{Size} & \textbf{BoN} & \textbf{Pair} & \textbf{$\tau$} \\
  \midrule
  \multicolumn{5}{@{}l}{\textit{Open-Source General VLMs}} \\
  InternVL3-8B     & 8B  & 40.4 & 71.6 & 0.466 \\
  Qwen2.5-VL-7B    & 7B  & 44.4 & 73.6 & 0.504 \\
  Qwen3-VL-4B      & 4B  & 47.9 & 75.8 & 0.538 \\
  Qwen3-VL-8B      & 8B  & 47.7 & \underline{77.2} & \underline{0.563} \\
  Qwen3-VL-32B     & 32B & \underline{49.9} & \textbf{77.8} & \textbf{0.592} \\
  Molmo2-4B        & 4B  & 43.2 & 71.9 & 0.485 \\
  Molmo2-8B        & 8B  & 42.6 & 70.8 & 0.476 \\
  InternVL3-78B    & 78B & \textbf{49.5} & 76.7 & 0.553 \\
  \midrule
  \multicolumn{5}{@{}l}{\textit{Open-Source Generative Reward Models}} \\
  R1-Reward        & 7B  & \textbf{44.8} & \textbf{68.4} & \textbf{0.496} \\
  MM-RLHF-Reward   & 7B  & 36.6 & 67.0 & 0.352 \\
  LLaVA-Critic     & 7B  & \underline{40.2} & 69.0 & \underline{0.421} \\
  \midrule
  \multicolumn{5}{@{}l}{\textit{Open-Source Discriminative Reward Models}} \\
  Skywork-VL-Reward & 7B & 46.7 & 74.4 & 0.499 \\
  IXC-2.5-Reward   & 7B  & \underline{48.7} & 74.0 & 0.491 \\
  \textbf{Molmo2-4B RM (Ours)} & \textbf{4B} & \textbf{50.7} & \textbf{77.4} & \textbf{0.550} \\
  \textbf{Qwen3-VL-4B RM (Ours)} & \textbf{4B} & 47.5 & \underline{73.8} & 0.482 \\
  \bottomrule
\end{tabular}
\end{center}
\caption{Full metrics on our \textbf{\videobench} (4-response, 495 samples).
\textbf{BoN} = best-of-N accuracy (\%, primary metric from Table~\ref{tab:main_results});
\textbf{Pair} = pairwise accuracy (\%);
\textbf{$\tau$} = Kendall's $\tau$ rank correlation.
Models: Skywork-VL-Reward~\citep{wang2025skyworkvlrewardeffectivereward}, IXC-2.5-Reward~\citep{zang2025internlmxcomposer25rewardsimpleeffectivemultimodal}, R1-Reward~\citep{zhang2025r1rewardtrainingmultimodalreward}, MM-RLHF-Reward~\citep{zhang2025mmrlhfstepforwardmultimodal}, LLaVA-Critic~\citep{xiong2025llavacriticlearningevaluatemultimodal}, Molmo2~\citep{clark2026molmo2openweightsdata}, Qwen3-VL~\citep{bai2025qwen3vltechnicalreport}.
Best per section in \textbf{bold}, second best \underline{underlined}.}\label{tab:video_full_metrics}
\end{table}

\subsection{Single-RM GRPO Hyperparameter Search}
\label{sec:grpo_single_rm}

To provide a rigorous comparison, we trained four single-RM GRPO policy variants covering two architectural choices and two learning-rate/KL configurations, all using the same base model, training data, and optimization setup as the multi-RM run.

\begin{itemize}
\item \textbf{Single-RM (LoRA-32)}: LoRA rank 32, learning rate $5{\times}10^{-6}$, KL coefficient 0.02. The most stable variant; reported in Table~\ref{tab:grpo_eval_standard}.
\item \textbf{Single-RM (LoRA-64)}: LoRA rank 64, learning rate $5{\times}10^{-6}$, KL coefficient 0.02. Exhibits reward hacking: the model degenerates to repeating exclamation marks on all inputs (VQAv2 $\approx$0\%).
\item \textbf{Single-RM (Full FT)}: Full fine-tuning, learning rate $5{\times}10^{-6}$, KL coefficient 0.02. Severe reward hacking: hallucination rate jumps to 52.1\% (from base 39.6\%).
\item \textbf{Single-RM (Full FT, lr1e6, KL0.5)}: Reduced learning rate $1{\times}10^{-6}$ and stronger KL penalty 0.5. Partially mitigates hacking (hallucination 38.5\%) but open-ended quality remains below base (WildVision 53.4 vs.\ base 54.6).
\end{itemize}

Two out of four configurations exhibit reward hacking, underscoring the instability of single-response absolute rewards under GRPO optimization. The multi-response RM, which provides a comparative reward signal, avoids this instability entirely.

\subsection{Training Data Details}
\label{app:training_data}

We curate training data from 881K raw samples across 10 source datasets, selecting 436K for the final training set (Table~\ref{tab:train_datasets_hf_used}). To balance dataset sizes, we weight each source proportionally to the square root of its size and upsample underrepresented task categories (e.g., reasoning, safety, document understanding). The multimodal portion draws from MM-RLHF, LLaVA-Critic, RLAIF-V, VLFeedback, POVID, and WildVision~\citep{zhang2025mmrlhfstepforwardmultimodal,xiong2025llavacriticlearningevaluatemultimodal,yu2025rlaifvopensourceaifeedback,li2024vlfeedbacklargescaleaifeedback,zhou2024aligningmodalitiesvisionlarge,lu2024wildvisionevaluatingvisionlanguagemodels}; the text portion from Tulu, Skywork, Nectar, and PKU-SafeRLHF~\citep{lambert2025tulu3pushingfrontiers,liu2024skyworkrewardbagtricksreward,starling2023,ji2025pkusaferlhfmultilevelsafetyalignment}.

\begin{table}[h!]
\begin{center}
\small
\setlength{\tabcolsep}{4pt}
\begin{tabular}{@{}lrrl@{}}
  \toprule
  Dataset & Collected & Used & $N$ \\
  \midrule
  \multicolumn{4}{@{}l}{\textbf{Multimodal}} \\
  \texttt{MMHAL/MM-RLHF}                                   & 16,321  & 16,321 & $N{=}3\sim5$ \\
  \texttt{lmms-lab/LLaVA-Critic-113k}                     & 71,331  & 56,635 & $N{=}2\sim13$ \\
  \texttt{openbmb/RLAIF-V-Dataset}                        & 83,132  & 64,097 & $N{=}2$ \\
  \texttt{MMInstruction/VLFeedback}                       & 80,258  & 63,448 & $N{=}4$ \\
  \texttt{YiyangAiLab/POVID\_preference\_data\_for\_VLLMs}& 17,184  & 17,184 & $N{=}2$ \\
  \texttt{WildVision/wildvision-battle}                   & 7,198   & 7,198  & $N{=}2$ \\
  \midrule
  \multicolumn{4}{@{}l}{\textbf{Text-only}} \\
  \texttt{allenai/llama-3.1-tulu-3-8b-preference-mixture} & 272,013 & 82,911 & $N{=}2$ \\
  \texttt{Skywork/Skywork-Reward-Preference-80K-v0.2}     & 77,004  & 45,487 & $N{=}2$ \\
  \texttt{berkeley-nest/Nectar}                           & 182,954 & 47,862 & $N{=}7$ \\
  \texttt{PKU-Alignment/PKU-SafeRLHF}                     & 73,870  & 35,292 & $N{=}2$ \\
  \midrule
  \textbf{Total} & \textbf{881,265} & \textbf{436,435} & \\
  \bottomrule
\end{tabular}
\end{center}
\caption{\textbf{Training data composition.} We curate 436K samples from 10 datasets, with 35\% containing $N{>}2$ responses for listwise training. $N$: number of responses per sample.}\label{tab:train_datasets_hf_used}
\end{table}

\subsection{GRPO Training Details}
\label{app:grpo_setup}

The GRPO policy uses full fine-tuning with a frozen vision tower for 500 steps. We use cosine learning rate decay with 10\% warmup and a minimum ratio of 0.2, learning rate $1 \times 10^{-5}$, batch size 8, and 2 PPO epochs per step. KL regularization (coefficient 0.05) is applied via the low-variance KL loss added to the policy gradient objective.

\subsection{\videobench{} Details}
\label{app:video_models}
\label{app:video_collection}

For each question, we generate responses from 19 diverse models spanning proprietary APIs (GPT-5~\citep{openai2025gpt5}, Claude Sonnet~4.5~\citep{anthropic2025claudesonnet45}, Gemini~2.5~Pro/Flash~\citep{comanici2025gemini}) and open-source models of varying scales (Molmo2-4B/8B~\citep{clark2026molmo2openweightsdata}, Qwen3-VL-4B/8B~\citep{bai2025qwen3vltechnicalreport}, InternVL3.5-4B/8B~\citep{zhu2025internvl3exploringadvancedtraining}, LLaVA-Video-7B~\citep{zhang2025llavavideovideoinstructiontuning}, MiniCPM-V4.5~\citep{yu2025minicpmv45cookingefficient}, Eagle2.5~\citep{chen2025eagle25boostinglongcontext}, VideoChat-Flash~\citep{li2025videochatflashhierarchicalcompressionlongcontext}, GLM-V4.1~\citep{glm2024chatglmfamilylargelanguage}, KeyEVL1.5~\citep{yang2025kwaikeyevl15technical}, PLM-3B/8B~\citep{cho2025perceptionlmopenaccessdatamodels}). Human annotators are presented with a video, a question, and two model responses side-by-side, and asked to select which response is better or declare a tie. In total, 1,116 crowdworkers produce approximately 94K pairwise judgments in a balanced tournament design, with each model pair compared roughly 551 times across the question set. The data was collected as part of the Molmo2 data collection effort~\citep{clark2026molmo2openweightsdata}.

\end{document}